\documentclass[journal]{IEEEtran}

\usepackage{graphics} 
\usepackage{epsfig} 
\usepackage{amsmath} 
\usepackage{amssymb}  
\usepackage[table]{xcolor}
\usepackage[ruled,vlined]{algorithm2e}

\usepackage{epsfig, graphicx, amsmath, amssymb, subfigure, color}

\begin{document}

\title{Sharable Clothoid-based Continuous Motion Planning for Connected Automated Vehicles}

\author{Sanghoon~Oh, Qi Chen, H.~Eric~Tseng, Gaurav Pandey, and~G\'abor~Orosz
\thanks{S. Oh and G. Orosz are with the Department of Mechanical Engineering, University of Michigan, Ann Arbor, MI 48109, USA
{\tt \{osh,orosz\}@umich.edu}}
\thanks{Q.~Chen, H.~E.~Tseng, and G.~Pandey are with Ford Motor Company, Dearborn, MI 48124, USA {\tt \{qchen51,htseng,gpandey2\}@ford.com}}    
\thanks{G.~Orosz is also with the Department of Civil and Environmental Engineering, University of Michigan, Ann Arbor, MI 48109, USA}
}

\markboth{}%
{Shell \MakeLowercase{\textit{et al.}}: Bare Demo of IEEEtran.cls for IEEE Journals}

\maketitle


\begin{abstract}
A continuous motion planning method for connected automated vehicles is considered for generating feasible trajectories in real-time using three consecutive clothoids. 
The proposed method reduces path planning to a small set of nonlinear algebraic equations such that the generated path can be efficiently checked for feasibility and collision. 
After path planning, velocity planning is executed while maintaining a parallel simple structure. Key strengths of this framework include its interpretability, shareability, and ability to specify boundary conditions.
Its interpretability and shareability stem from the succinct representation of the resulting local motion plan using a handful of physically meaningful parameters. Vehicles may share these parameters via V2X communication so that the recipients can precisely reconstruct the planned trajectory of the senders and respond accordingly. 
The proposed local planner guarantees the satisfaction of boundary conditions, thus ensuring seamless integration with a wide array of higher-level global motion planners. 
The tunable nature of the method enables tailoring the local plans to specific maneuvers like turns at intersections, lane changes, and U-turns.

\end{abstract}

\begin{IEEEkeywords}
local motion planning, G2 smoothness, connected automated vehicles, V2X communication.
\end{IEEEkeywords}

\IEEEpeerreviewmaketitle

\section{Introduction}

\IEEEPARstart{A}{tonomous} driving involves various levels of planning including route planning, behavioral planning, and motion planning \cite{ersal2020connected,wubing2022}. 
The latter specifies the desired trajectory of the vehicle which should meet multiple requirements -- feasibility, smoothness, and free of collisions. 
However, optimal motion planning subject to geometric and kinematic constraints is PSPACE-hard \cite{reif1979complexity}. 
Motion primitives can dramatically decrease the complexity of the planning problem, but they should provide flexibility and ease of tuning to generate smooth feasible trajectories. 
This paper proposes a local motion planning method, akin to a motion primitive. 
Interpretable parametrization promotes sherability via V2X communication, while precise alignment with boundary conditions supports integration with higher-level planners. 
In addition, tunability, arising from adjustable parameters, enhances the versatility. 
These core strengths are applied to both path planning and velocity planning.

There have been various approximate approaches for motion planning of connected automated vehicles (CAVs) using variational methods \cite{bertsekas1997nonlinear}, graph-search methods \cite{karaman2011sampling,schmerling2015optimal}, and incremental search methods \cite{webb2013kinodynamic,karaman2013sampling}. 
These planning methods are well suited for complex scenarios where obstacles are distributed over an unstructured environment \cite{lavalle2006planning} and no specifications of the local waypoints are given. 
However, they typically yield local optimality and are subject to heavy computational burden. 
Considering that urban roads are structured environments, where a set of waypoints can be pre-defined along the route \cite{gonzalez2016review}, we propose interpolating curve planning to address the above challenges and efficiently generate smooth local trajectories between given waypoints. 

Interpolating curve planning methods may use different kinds of base curves, and clothoids are among the most intriguing ones as they can make smooth transitions between straight lines and curved arcs \cite{gonzalez2016review,fleury1995primitives,kellynagy2003,botros2022tunable}. 
In a recent work \cite{bertolazzi2018g2}, $G^2$ Hermite interpolation was proposed to generate paths that consist of three consecutive clothoids, are second-order smooth, and satisfy the boundary conditions for arbitrary initial and final positions, angles, and curvatures. 
Within this paper, we extend this method by encompassing feasibility constraints and by augmenting the path planning with a compatible velocity plan.
These extensions enable us to generate local motion plans, between waypoints cascaded down from a higher-level planner, for CAVs traversing in structured environments.

Wireless vehicle-to-everything (V2X) communication can be utilized to improve the safety of automated vehicles, since it can provide information about other connected agents, including those beyond the line of sight. 
Use cases include control systems that can enhance stability \cite{orosz2016connected,jin2018connected,naus2010string}, energy efficiency \cite{he2019fuel}, and safety \cite{liu2020freeway,shen2021saving}, mostly focusing on the longitudinal dynamics of vehicles. 
Current messages, like the basic safety message (BSM), only include the current position and velocity of the vehicle. 
There have been recent efforts to enhance the utility of messages by including additional useful information such as intent \cite{wang2023intent1,wang2023intent2,van2021cooperative}. 
It was also shown that connectivity can enhance coordination in complicated scenarios like unsignalized intersections \cite{hult2020optimisation,chalaki2021priority}. 
To enhance traffic safety and mobility with minimal extension of messages, this paper proposes local motion planning that can be encoded in a compact V2X message.
It will be demonstrated that such motion plans, including both path and velocity profiles, can be easily encoded and decoded, and can be utilized to resolve conflicts in scenarios when multiple agents share their local plans via V2X communication.

\subsection{Related work}

For mobile robots with nonholonomic constraints, there have been efforts to provide suitable local motion plans based on optimization methods \cite{webb2013kinodynamic, blanco2015tp}. 
For example, Dubins’ path, which is shown to be optimal in terms of path length \cite{dubins1957curves}, is used as the local motion planning method for kinodynamic RRT* \cite{karaman2013sampling}. 
However, such plans cannot provide curvature continuity and are not suitable for an actual vehicle because the desired steering angle profile consists of only maximum steering and zero steering. 
One may tackle this problem by solving an optimization problem based on a nonlinear model with continuous curvature \cite{hwan2013optimal}, but applying nonlinear optimization for every extension process of an RRT leads to heavy computational load. 
Path smoothing is another popular approach that can lead to a continuous curvature profile \cite{elbanhawi2015continuous,simba2016real}, but such methods  typically modify the boundary conditions. 
There also exist reinforcement learning-based motion planning approaches \cite{gomez2012optimal,sivashangaran2021intelligent} that require maintaining a Q-table for all possible state-action pairs, which cannot be encoded into shareable V2X messages \cite{aradi2020survey}. 

Interpolating spline methods may be used to solve the boundary value problem (BVP) of local motion planning. 
There are different interpolating methods with different primitives, including polynomial splines \cite{piazzi2000quintic,ghilardelli2013path}, Bezier curve \cite{qian2016motion,elhoseny2018bezier}, B-spline \cite{elbanhawi2015continuous,huang2021personalized}, and clothoids \cite{sharma2019survey}. 
For example, the local path planner in \cite{ghilardelli2013path} results in smooth curves using polynomial splines (parameterized with 8 free parameters), while the optimization-based local motion planner in \cite{huang2021personalized} uses $G^2$ smooth primitives. 
However, finding the parameters of splines usually requires solving nonlinear non-convex optimization problems, which is difficult to apply in real time. 
Clothoids can be used for path design since the curvature, and thus the tangent of the desired steering angle, are affine functions of the arclength \cite{fraichard2004reeds,wilde2009computing,tian2021continuous}. 
This is why clothoids are also used for designing road curvature profiles \cite{marzbani2015better}. 

A single clothoid might not be enough to solve the BVP, but several consecutive clothoids are. 
In \cite{banzhaf2018g}, a $G^3$ smooth planning method is proposed, but it is limited to scenarios when zero curvatures at both boundaries are enforced, and therefore it is only appropriate for parking lot maneuvers. 
In \cite{bertolazzi2018g2}, a $G^2$ Hermite interpolation method is proposed that solves the BVP for given positions, yaw angles, and curvatures at the boundaries using three consecutive clothoids. 
It is shown that this leads to a system of algebraic equations that can be numerically solved for arbitrary boundary conditions. 
This work motivates our study to create a feasible local planner for connected automated vehicles when curvature limits are introduced by the steering system.

\subsection{Contributions of the paper}

Our three-clothoid motion planning method first generates a $G^2$ continuous path (i.e., a path with continuous curvature) and subsequently establishes a smooth velocity profile, known as path velocity decomposition (PVD). 
Thanks to the $G^2$ continuity, the exact swept volume of the vehicle can be obtained by tracing a few points of the vehicle's body. 
We exploit this when developing a novel collision-checking method that outperforms sampling-based collision-checking methods in terms of consistency. 
We also utilize the structure of the path plan for velocity planning by designing a constant acceleration for each clothoid and then generating a piece-wise smooth velocity plan 
using constant jerk smoothing.
This way we obtain a compact representation of the motion plan, represented by a few parameters of clear physical interpretation.

There are three significant advantages of the proposed framework. 
First, it generates an interpretable continuous local motion plan which can be summarized and reproduced with only a few parameters of clear physical meanings. 
This small number of parameters makes it easy to encode the plan and share it via V2X messages.  
Second, the planned motion satisfies the boundary condition exactly and does not require an additional smoothing.
This makes it easy to integrate the local motion plans with different higher-level motion planners.
Third, the method possesses a tunability feature which enables one to generate multiple local plans that satisfy the same boundary condition. 
Then the user may select the plan based on different objectives such as travel time and/or maximum curvature. 
We will demonstrate that the method is applicable to a large variety of maneuvers, including turns at intersections, lane changes, and U-turns, and that it can be utilized to resolve conflicts between connected automated vehicles. 

The rest of the paper is organized as follows.
We formulate the problem of motion planning in Section~\ref{sec:problem_statement}. 
In Section~\ref{sec:path} the path planning problem is described followed by the velocity planning in Section~\ref{sec:velocity}. 
In Section~\ref{sec:implementation} local planning examples are compared to benchmark planning algorithms for representative urban maneuvers, the encoding/decoding of the plans to/from V2X messages is explained, and the conflict resolution between multiple CAVs is discussed. 
We conclude the results in Section~\ref{sec:conclusion} where we also lay out some future research directions.

\section{Problem Statement}\label{sec:problem_statement}

Below we formulate the motion planning problem in a mathematical form. It assumed that the initial and the final waypoints are given by a higher-level planner. 
First, we describe the mathematical model used for planning, followed by formulating the boundary value problem one needs to solve to plan a trajectory.

\subsection{Single track vehicle model}

\begin{figure}[!t]

\begin{center}
\begin{picture}(180,190)
\put(0.0, 0.0){\epsfig{file=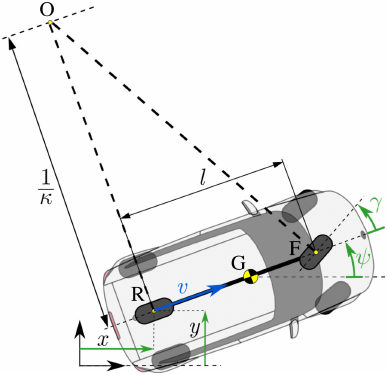, width=65.0mm}}   
\put(62.0,35.0){R} 
\put(110.0,50.0){G} 
\put(138.0,57.0){F} 
\put(18.0,172.0){O} 
\end{picture}
\end{center}

\caption{Single track model of an automobile used in the paper.}
\label{fig:kinematic_bicycle}
\vspace{-4mm}
\end{figure}
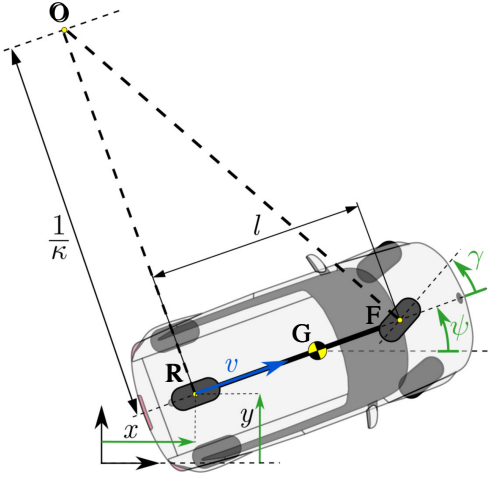

In this paper, a single-track model shown in Fig.~\ref{fig:kinematic_bicycle} is used for motion planning. 
For simplicity, the front wheels are merged into one wheel and the rear wheels are merged into one wheel. 
No slip conditions at the front and the rear are assumed which are good approximations for most urban scenarios \cite{kong2015kinematic,mi2023UAV}. 
The vehicle's motion is described by three configuration coordinates: $x$ and $y$ denote the position of the center R of the rear axle and $\psi$ denotes the yaw angle. 
In addition, $v$ denotes the longitudinal velocity of the body which is assumed to be non-negative. 
Thus, the state vector consists of the four states ${[x,y,\psi,v]^\top}$ defined in the state space ${\mathcal{X} = \mathbb{R}^2 \times \mathbb{S} \times \mathbb{R}_{\geq 0}}$. 
The input vector ${[\gamma,a]^\top}$ consists of the front steering angle $\gamma$ and the longitudinal acceleration $a$ defined in the input space ${\mathcal{U} = \mathbb{S} \times \mathbb{R}}$. 
The nonlinear differential equation is formulated as
\begin{equation}\label{eq:bicycle_model}
\begin{split}
\dot{x} &= v \cos{\psi},
\\
\dot{y} &= v \sin{\psi},
\\
\dot{\psi} &= \dfrac{v}{l} \tan{\gamma},
\\
\dot{v} &= a,
\end{split}
\end{equation}
where the dot represents the derivative with respect to time $t$ and $l$ denotes the wheelbase, i.e., the distance between the center of the rear axle R and the center of the front axle F; see Fig.~\ref{fig:kinematic_bicycle}. 

The input constraints are expressed as
\begin{equation}\label{eq:constraint}
\begin{split}
|\gamma| &\leq \gamma_{\textrm{max}},
\\ 
|\dot{\gamma}| &\leq \Omega_{\textrm{max}},
\\
a_{\textrm{min}} &\leq a \leq a_{\textrm{max}},
\\
|\dot{a}| &\leq j_{\textrm{max}}.
\end{split}   
\end{equation}
Here we use the values ${\gamma_{\textrm{max}}=\pi/6}$, ${\Omega_{\textrm{max}}=2\pi~\textrm{1/s}}$, ${a_{\textrm{min}}=-8~\textrm{m/s}^2}$, ${a_{\textrm{max}}=3~\textrm{m/s}^2}$, and ${j_{\textrm{max}}=2~\textrm{m/s}^3}$. Note that the acceleration constraint is related to the limits of the powertrain and the brake system of the vehicle. The jerk limit is considered for the ride comfort of the passengers. 
An additional state constraint, which accounts for the lateral acceleration limit and a rule of the road, will be introduced in Section~\ref{sec:velocity}. 

From the no-slip conditions of the wheels, the instantaneous curvature of the rear axle center can be expressed by
\begin{equation}\label{eq:curvature}
    \kappa=\frac{\tan{\gamma}}{l}.
\end{equation}
This shows that setting the steering angle in \eqref{eq:bicycle_model} is equivalent to setting the curvature of the path. Additionally, the steering input constraint in \eqref{eq:constraint} can be translated to a curvature constraint.

\subsection{Motion planning problem}

To define the motion planning problem, we first define the workspace, configuration space, and state space precisely. 
The workspace is the ${(x,y)}$ plane, that is, ${\mathcal{W}=\mathbb{R}^2}$. 
Part of this plane is occupied by the vehicle itself while other parts may be occupied by obstacles.
The configuration space is spanned by the scalar coordinates that are needed to unambiguously describe the configuration of the vehicle. 
These are the $x$ and $y$ positions of the rear axle center point R and the yaw angle $\psi$, that is, the configuration space has the topological structure ${\mathcal{C}=\mathbb{R}^2 \times \mathbb{S}}$.
The state space includes all variables needed to describe the state of the vehicle. 
In our case the longitudinal speed $v$ is added to the configuration coordinates yielding the state space ${\mathcal{X}=\mathbb{R}^2 \times \mathbb{S} \times \mathbb{R}_{\geq 0}}$. 
Note that the lateral velocity and the yaw rate are not used as state variables since we assume that neither the front nor the rear wheels have side slip \cite{wubing2022}.
We assume that a higher-level planner defines the desired waypoints in the configuration space $\mathcal{C}$ and then focus on planning the motion in the state space $\mathcal{X}$.

Given the initial and final configurations, the "relative configuration" between them can be found by putting the initial configuration at the origin of the configuration space and changing the basis accordingly. 
The transformed final configuration is denoted as ${[\Delta x, \Delta y, \Delta \psi ]^{\top}}$. 
Thus, the local motion plan needs to satisfy the boundary conditions
\begin{equation}\label{eq:Problem_1}
\begin{split}
    x(0) &=0,\qquad x(T)=\Delta x, 
    \\
    y(0) &=0,\qquad y(T)=\Delta y, 
    \\
    \psi(0) &=0,\qquad \psi(T)=\Delta\psi, 
    \\
    v(0) &=v_\textrm{0},\qquad v(T)=v_\textrm{f},
    \\
    \gamma(0) &= \gamma_{\textrm{0}},\qquad \gamma(T)=\gamma_{\textrm{f}}.
\end{split}  
\end{equation}

Our aim with the local motion planner is to prioritize interpretability, shareability, and the ability to set boundary conditions (e.g., $v_0$ and $v_\textrm{f}$) rather than focusing on generating a trajectory with the minimum travel time. 
Here we assume that the initial velocity $v_0$ is given by the global planner while the final velocity $v_\textrm{f}$ and the travel time $T$ are not fixed. 
In general, the global planner can specify either $v_0$ or $v_\textrm{f}$.
The initial steering angle $\gamma_0$ and the final steering angle $\gamma_{\textrm{f}}$ are also assumed to be given by the global planner. 
The motion planning problem with arbitrary final steering angle will be considered in future studies. 
It is also assumed that the location and shape of stationary obstacles are known. 
Consequently, the local motion planning problem can be formulated as follows.

\noindent \textbf{Problem 1.}---Trajectory Planning. Given the initial and final positions and headings, and the initial steering angle, find a trajectory (which includes path and speed profile) that satisfies the differential equations \eqref{eq:bicycle_model}, the constraints \eqref{eq:constraint}, and the boundary conditions \eqref{eq:Problem_1}.

To solve this problem the local motion planning algorithm returns a path of continuous curvature and velocity which is parameterized by the traveled distance. 
Note that by limiting the steering rate and jerk in \eqref{eq:constraint}, the solution of \eqref{eq:bicycle_model} is continuously differentiable which also leads to the continuity of the curvature. 
The motion planning algorithm is divided into a separate path planner and velocity planner, which are described in the following sections.

\section{Path Planing}\label{sec:path}

To solve Problem 1, we develop a local motion planner which first designs the geometric path and then the velocity profile. 
To be independent of the velocity profile, the geometric path is represented as a function of the distance $s$ traveled by the center or the rear axle. 
This is often referred to as arclength parameterization. 
That is, the trajectory is expressed as ${[x(s), y(s), \psi(s), v(s)]^\top}$ in the state space ${\mathbb{R}^2 \times \mathbb{S} \times \mathbb{R}_{\geq 0}}$.

The derivatives of the configuration variables ${x,y,\psi}$ with respect to the arclength $s$ are denoted by ${x',y',\psi'}$. 
These can be obtained using the chain rule:
\begin{equation}
    \dot{x} = \frac{\textrm{d}x}{\textrm{d}t} = \frac{\textrm{d}x}{\textrm{d}s} \frac{\textrm{d}s}{\textrm{d}t} = x'\dot{s} = x'v.
\end{equation}
Thus, \eqref{eq:bicycle_model} and \eqref{eq:curvature} yield the differential equations
\begin{equation}\label{eq:G2_problem}
\begin{split}
    x'&=\cos{\psi}, 
    \\
    y'&=\sin{\psi},
    \\
    \psi'&=\kappa,
\end{split}  
\end{equation}
while \eqref{eq:Problem_1} yields the boundary conditions
\begin{equation}\label{eq:G2_problem_BC}
\begin{split}
    x(0)&=0,\qquad x(s_\textrm{f})=\Delta x, 
    \\
    y(0)&=0,\qquad y(s_\textrm{f})=\Delta y, 
    \\
    \psi(0)&=0,\qquad \psi(s_\textrm{f})=\Delta\psi, 
    \\
    \kappa(0)&=\kappa_0,\quad \,\,\, \kappa(s_\textrm{f})=\kappa_2. 
\end{split}  
\end{equation}
Here $s_\textrm{f}$ denotes the total distance traveled which is not fixed a priori. 
The initial and final curvatures $\kappa_0$ and $\kappa_2$ correspond to the initial and final steering angles $\gamma_0$ and $\gamma_{\textrm{f}}$, respectively. 
Finally, using \eqref{eq:curvature}, the first constraint in 
\eqref{eq:constraint} can be rewritten for the curvature:
\begin{equation}\label{eq:constraint2}
\left|\kappa(s)\right| \leq \frac{\tan{\gamma_{\textrm{max}}}}{l}.
\end{equation}
Thus, the path planning can be formally cast as follows.

\noindent \textbf{Problem 2.}---Path Planning. Given the differential equation \eqref{eq:G2_problem} and the constraint \eqref{eq:constraint2}, find a continuous function $\kappa(s)$ that satisfies the boundary conditions \eqref{eq:G2_problem_BC}.

\noindent Notice that we require the continuity of the function $\kappa(s)$ to ensure $G^2$ continuity of the path. 

We solve Problem 2 using a three-clothoid $G^2$ Hermite interpolation method \cite{bertolazzi2018g2}. 
We emphasize that none of the existing methods, including polynomial-based spline, B-spline, and Dubins path, meet all four of the following criteria: (i) ensuring $G^2$ continuity; (ii) exactly satisfying the boundary conditions; (iii) encoding the path with a fixed number of parameters which have a clear physical interpretation; (iv) having tunable parameters which allow the generation of multiple paths for given boundary conditions. With the three-clothoid method to be described below, a family of local paths, each having 2 tunable parameters can be generated by solving 8 algebraic nonlinear equations with 10 unknowns. 
These equations can be solved numerically such that the resulting path exactly meets the boundary conditions.

\subsection{Three-clothoid path generation}

By integrating \eqref{eq:G2_problem}, the end boundary conditions \eqref{eq:G2_problem_BC} become
\begin{equation}\label{eq:G2_problem_int}
\begin{split}
\Delta x &= \int_0^{s_\textrm{f}} \cos\Big(\textstyle\int_0^s \kappa(\sigma) \textrm{d} \sigma \Big) \textrm{d} s,
\\
\Delta y &= \int_0^{s_\textrm{f}} \sin\Big(\textstyle\int_0^s \kappa(\sigma) \textrm{d} \sigma \Big) \textrm{d} s,
\\
\Delta \psi & = \int_0^{s_\textrm{f}} \kappa(s) \textrm{d} s.
\end{split}
\end{equation}
Thus, assuming a particular function $\kappa(s)$ for the curvature determines the path, and the simpler the algebraic form is the easier to evaluate the integrals in \eqref{eq:G2_problem_int}.
Here use piecewise linear functions for $\kappa(s)$ which makes its integrals piecewise quadratic and results in a piece-wise three-clothoid path.

\begin{figure}
\centering
\includegraphics[width=3.2 in]{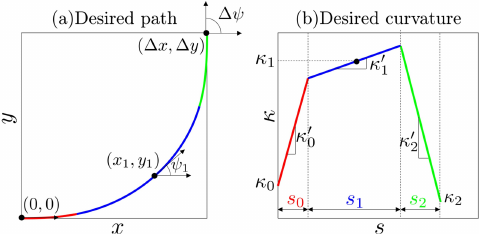}
\caption{Three-clothoid path plan for a left turn with relevant variables depicted. 
The curvature is continuous and all boundary conditions are met.}
\label{fig:clothoids_path_ex}
\vspace{-4mm}
\end{figure}

Fig.~\ref{fig:clothoids_path_ex} depicts an example three-clothoid path. 
The initial configuration is the origin ${(0,0,0)}$, while the initial curvature is $\kappa_0$. 
The final configuration is ${(\Delta x, \Delta y, \Delta \psi)}$ and the final curvature is $\kappa_2$. 
Moreover, $x_1$, $y_1$ and $\psi_1$ denote the location and orientation of the middle point of the second clothoid, and ${\kappa_1=\kappa\big(s_0+\frac{s_1}{2}\big)}$ gives the curvature at this middle point. 
Finally, $\kappa'_0$, $\kappa'_1$ and $\kappa'_2$ denote the constant sharpness of the first, second, and third clothoids, respectively. 
The corresponding continuous curvature function $\kappa(s)$ is expressed as 
\begin{equation}\label{eq:curv}
\begin{split}
    &\texttt{path}(s_0,s_1,s_2,\kappa_0,\kappa_1,\kappa_2,\kappa'_0,\kappa'_1,\kappa'_2) 
    \\
    &=
\begin{cases}
    \kappa_0+\kappa'_0 s, \quad &\text{if } 0 \leq s \leq s_0,
    \\
    \kappa_1 + \kappa'_1(s-s_0-\frac{s_1}{2}), \quad &\text{if } s_0 \leq s \leq s_0+s_1,
    \\
    \kappa_2 + \kappa'_2 (s-s_\textrm{f}), \quad &\text{if } s_0+s_1 \leq s \leq s_\textrm{f},
\end{cases}
\end{split}
\end{equation}
where ${s_\textrm{f}=s_0+s_1+s_2}$. 
The existence of an unconstrained solution of such a local path is proven numerically in \cite{bertolazzi2018g2} with a systematic selection of parameters ${s_0,s_2>0}$. 
We also remark that out of the 9 parameters listed in \eqref{eq:curv}, only 7 are independent. 
For example, one may express the sharpnesses $\kappa'_0$ and $\kappa'_2$ from the continuity equations
\begin{equation}\label{eq:curvature_continuity}
\begin{split}
\kappa_1 - \kappa'_1 \frac{s_1}{2} &= \kappa_0 + \kappa'_0 s_0,
\\
\kappa_1 + \kappa_1' \frac{s_1}{2} &= \kappa_2 - \kappa'_2 s_2.
\end{split}
\end{equation}

To express the position of a clothoid path (cf.~\eqref{eq:G2_problem_int}) let us define the Fresnel integrals
\begin{equation}\label{eq:Fresnel}
\begin{split}
    C(a,b,c) &:=\int_{0}^{1}\cos\Big(\frac{a}{2}\sigma^2+b\sigma+c\Big)\textrm{d}\sigma, 
    \\
    S(a,b,c) &:=\int_{0}^{1}\sin\Big(\frac{a}{2}\sigma^2+b\sigma+c\Big)\textrm{d}\sigma.
\end{split}  
\end{equation}
Using these one can set up the continuity conditions and the boundary conditions for three clothoids yielding 8 nonlinear algebraic equations with 15 variables:
\begin{eqnarray}\label{eq:Clothoids_problem}
    x_1-\frac{s_1}{2} C\Big(\frac{\kappa'_1 s_1^2}{4},-\frac{\kappa_1s_1}{2},\psi_1\Big) \!\!\!\!\!&=&\!\!\!\!\! s_0 C\big(\kappa'_0 s_0^2,\kappa_0s_0,0\big), 
    \\
    x_1+\frac{s_1}{2} C\Big(\frac{\kappa'_1 s_1^2}{4},\frac{\kappa_1s_1}{2},\psi_1\Big) \!\!\!\!\!&=&\!\!\!\!\! \Delta x-s_2 C(\kappa'_2 s_2^2,-\kappa_2 s_2,\Delta\psi),\nonumber
    \\
    y_1-\frac{s_1}{2} S\Big(\frac{\kappa'_1 s_1^2}{4},-\frac{\kappa_1s_1}{2},\psi_1\Big) \!\!\!\!\!&=&\!\!\!\!\! s_0 S\big(\kappa'_0 s_0^2,\kappa_0s_0,0\big), \nonumber
    \\
    y_1+\frac{s_1}{2}S\Big(\frac{\kappa'_1 s_1^2}{4},\frac{\kappa_1s_1}{2},\psi_1\Big) \!\!\!\!\!&=&\!\!\!\!\! \Delta y-s_2S\big(\kappa'_2 s_2^2,-\kappa_2 s_2,\Delta\psi\big),\nonumber
    \\
    \psi_1-\frac{\kappa_1s_1}{2}+\frac{\kappa_1's_1^2}{8} \!\!\!\!\!&=&\!\!\!\!\! \kappa_0s_0 + \frac{\kappa'_0s_0^2}{2},\nonumber
    \\
    \psi_1+\frac{\kappa_1s_1}{2}+\frac{\kappa_1's_1^2}{8} \!\!\!\!\!&=&\!\!\!\!\! \Delta\psi + \frac{\kappa'_2s_2^2}{2},\nonumber
    \\
    \kappa_1-\frac{\kappa_1's_1}{2}\!\!\!\!\!&=&\!\!\!\!\! \kappa_0+\kappa'_0s_0, \nonumber
    \\
    \kappa_1+\frac{\kappa_1's_1}{2} \!\!\!\!\!&=&\!\!\!\!\! \kappa_2-\kappa'_2s_2. \nonumber
\end{eqnarray}

\begin{figure}[!t]
\center
\includegraphics[width=3.3 in]{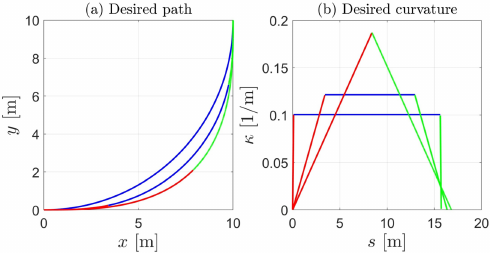}
\caption{Multiple paths generated for the same boundary conditions by tuning the arclength $s_0$ of the first clothoid and the arclength $s_2$ of the last clothoid.}
\label{fig:multiple_paths}
\vspace{-4mm}
\end{figure}

Among the 15 variables, ${(\Delta x, \Delta y, \Delta \psi, \kappa_0, \kappa_2)}$ are given. 
The variables ${(s_0, s_2)}$ are selected by us, and \eqref{eq:Clothoids_problem} is solved for the 8 remaining unknowns ${(x_1, y_1, \psi_1, \kappa_1, \kappa'_0, \kappa'_1, \kappa'_2, s_1)}$. 
Using algebraic manipulations, \eqref{eq:Clothoids_problem} can be reduced to two nonlinear equations for the two unknowns $s_1$ and $\kappa'_1$ \cite{bertolazzi2018g2}. 
These can be solved using Newton's method. The other unknowns can be expressed as a function of $s_1$ and $\kappa'_1$. 
We remark that for given ${(\Delta x, \Delta y, \Delta \psi, \kappa_0, \kappa_2)}$ multiple paths can be generated by tuning the parameters $s_0$ and $s_2$. 
All of these plans satisfy the boundary conditions. 
Such tunability is demonstrated for a left turn in Fig.~\ref{fig:multiple_paths} for the special case of ${s_0=s_2}$.  

\subsection{Online feasibility validation}

For some boundary conditions, there may not exist a feasible path. 
In order to check this, we establish feasibility charts that can be generated offline. 
A feasibility chart depicts the region in the configuration space where the solution of \eqref{eq:Clothoids_problem} leads to a curvature profile ${\kappa(s), s \in [0,s_{\rm f}]}$ that meets the inequality constraint \eqref{eq:constraint2}. 
Since the three-clothoid path depends on the free parameters ${s_0, s_2}$, the feasible region differs for different selections of these parameters. 

Here, to reduce the search space, we restrict ourselves to the case ${s_0=s_2}$ and also set ${\kappa_2=0}$.
That is, we vary the variables ${(\Delta x, \Delta y, \Delta \psi, \kappa_0, s_0)}$ in \eqref{eq:Clothoids_problem} to find the boundary where the equality
\begin{equation}\label{eq:constraint3}
\left|\kappa(s)\right| = \frac{\tan{\gamma_{\textrm{max}}}}{l},
\end{equation}
holds.
To efficiently solve \eqref{eq:Clothoids_problem}, \eqref{eq:constraint3}, the bisection method \cite{bachrathy2012bisection} is used. 
The resultant solution is a four-dimensional boundary in the five-dimensional space ${\mathbb{R}^2 \times \mathbb{S} \times \mathbb{R} \times \mathbb{R}_{\geq 0}}$ spanned by the variables ${(\Delta x, \Delta y, \Delta \psi, \kappa_0, s_0)}$. 
For effective visualization, we fix ${(\Delta \psi, \kappa_0, s_0)}$ and depict the 1-dimensional boundary in the 2-dimensional space ${(\Delta x, \Delta y)}$. 

\begin{figure}[!t]
\center
\includegraphics[width=3.3 in]{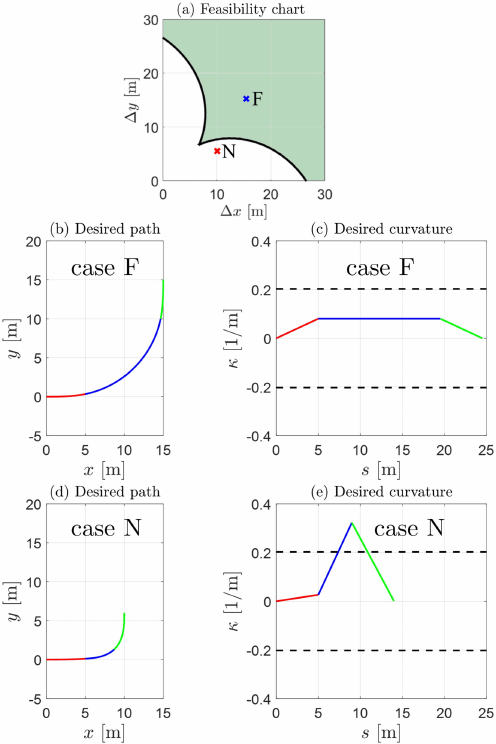}
\caption{(a) Feasibility chart in the ${(\Delta x, \Delta y)}$ plane where green shading indicates the feasible region. 
(b)-(c) Example for a feasible path. 
(d)-(e) Example for an infeasible path. 
The values ${\Delta \psi = \frac{\pi}{2}, \kappa_0 = \kappa_2 =0, s_0 = s_2 = 5\, {\rm m}}$ are fixed.}
\label{fig:feasibility_description}
\vspace{-4mm}
\end{figure}

\begin{figure*}[!t]
\center
  \includegraphics[width=4.8 in]{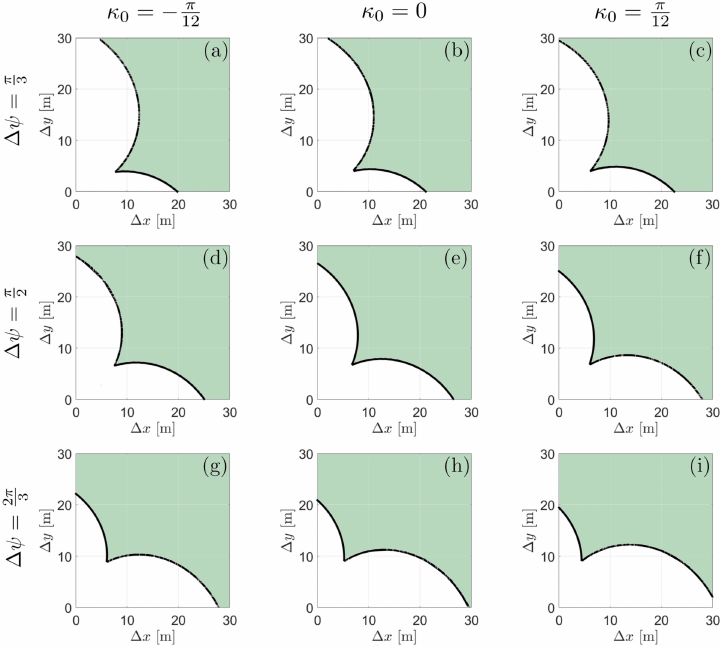}

  \caption{Feasibility charts in the ${(\Delta x, \Delta y)}$ plane where green shading indicates the feasible regions. 
  The values of ${\Delta \psi, \kappa_{0}}$ differ for each panel as indicated while the values ${s_0 = s_2 = 5\, {\rm m}}$ ${\kappa_2=0}$ are fixed.}
  \label{fig:feasibility_examples}
  \vspace{-4mm}
\end{figure*}

Fig.~\ref{fig:feasibility_description}(a) shows a feasibility chart where the feasible domain is shaded green. 
The obtained path and the corresponding curvature profile for the feasible point F are shown in Fig.~\ref{fig:feasibility_description}(b)-(c), while the path and curvature profile for the infeasible point N are shown in Fig.~\ref{fig:feasibility_description}(d)-(e).  
Notice that in the latter case, the curvature exceeds the constraint indicated by the upper dashed horizontal line.
In Fig.~\ref{fig:feasibility_examples} we show feasibility charts for different values of $\Delta \psi$ and $\kappa_0$ while still using ${s_0 = s_2}$ and  ${\kappa_2=0}$.
Notice that the charts look qualitatively similar but they become asymmetric once ${\Delta \psi \neq \frac{\pi}{2}}$ and ${\kappa_0\neq 0}$.

\section{Velocity Planing}\label{sec:velocity}

For each path a longitudinal velocity profile shall be generated. 
The last equation in \eqref{eq:bicycle_model} combined with ${\dot s = v}$ yields 
\begin{equation}\label{eq:longitudinal_problem}
\begin{split}
    &\dot{s}=v, 
    \\
    &\dot{v}=a. 
\end{split}  
\end{equation}
This is augmented by the boundary conditions 
\begin{equation}\label{eq:longitudinal_problem2}
\begin{split}
    s(0)&=0,\quad\,\,\, s(T)=s_\textrm{f},
    \\
    v(0) &=v_0,\quad v(T)=v_\textrm{f},
\end{split}  
\end{equation}
cf.~the fourth row in \eqref{eq:Problem_1}. 
As mentioned above, $v_\textrm{f}$ is unknown a priori. Also, while $s_\textrm{f}$ is obtained from the path plan, $T$ is not yet determined. 
The solution of the boundary value problem  \eqref{eq:longitudinal_problem},\eqref{eq:longitudinal_problem2} provides $s(t)$ and $v(t)$. The inverse of $s(t)$ gives time as a function of arclength $t(s)$, which can be substituted into $v(t)$ to get velocity plan as a function of arclength $v(s)$. Then the arclength parameterized acceleration is obtained as
\begin{equation}\label{eq:longitudinal_arclength}
    a(s) = \frac{\textrm{d}v}{\textrm{d}t} = \frac{\textrm{d}v}{\textrm{d}s} \frac{\textrm{d}s}{\textrm{d}t} = \frac{\textrm{d}v(s)}{\textrm{d}s} v(s).
\end{equation}
Also, we use the last two equations of \eqref{eq:constraint} and with an additional constraint on the lateral acceleration when generating the longitudinal plan. 
These result in 
\begin{equation}\label{eq:longitudinal_problem3}
\begin{split}
    &a_{\textrm{min}} \leq a \leq a_{\textrm{max}}, 
    \\
    &|\dot{a}| \leq j_{\textrm{max}}.
    \\
    &0 \leq v(s) \leq \overline{v}(s),
\end{split}  
\end{equation}
where $\overline{v}(s)$ encodes the steering rate constraint and the lateral acceleration constraint as derived below. 

Given the curvature plan $\kappa(s)$, one may limit the lateral acceleration at point R by limiting the velocity according to
\begin{equation}\label{eq:ride_comfort_constraint}
    |\kappa(s)|v^2(s) \leq a_{\textrm{lat},\textrm{max}},
\end{equation}
see \cite{wubing2022}. 
Here we use the value ${a_{\textrm{lat},\textrm{max}}=3~\textrm{m/s}^2}$.
Moreover, the steering rate constraint in \eqref{eq:constraint} can be translated to a velocity constraint taking the time derivative of \eqref{eq:curvature}:
\begin{equation}
\dot{\gamma} = l\, v(s)\, \kappa'(s)\cos^2\gamma = \frac{l\, v(s)\, \kappa'(s)}{1+\tan^2\gamma} .
\end{equation}
Substituting \eqref{eq:curvature} results in
\begin{equation}
\dot{\gamma} =  \frac{l\, v(s)\, \kappa'(s)}{1+l^2\kappa^2(s)},
\end{equation}
which yield
\begin{equation}\label{eq:steering_rate_constraint}
\left|\frac{l v(s)\, \kappa'(s)}{1+l^2\kappa^2(s)}\right| \leq \Omega_{\textrm{max}}.
\end{equation}
The constraints \eqref{eq:ride_comfort_constraint} and \eqref{eq:steering_rate_constraint} can be incorporated in the third row of \eqref{eq:longitudinal_problem3} as
\begin{equation}\label{eq:velocity_constraint}
    \overline{v}(s) = \min\bigg\{\sqrt{\frac{a_{\textrm{lat},\textrm{max}}}{|\kappa(s)|}},\, \Omega_{\textrm{max}} \frac{1+l^2\kappa^2(s)}{l\, |\kappa'(s)|}\bigg\}.
\end{equation}
Thus, the velocity planning problem is formulated as follows. 

\noindent \textbf{Problem 3.}---Velocity Planning. Given a continuous path composed of three connected clothoids with prescribed curvature ${\kappa(s), s\in [0 , s_\textrm{f}]}$, find a velocity profile $v(s)$ that maintains constant acceleration within each clothoid segment, subject to the dynamics \eqref{eq:longitudinal_problem}, the boundary condition \eqref{eq:longitudinal_problem2}, and the constraints \eqref{eq:longitudinal_problem3} and \eqref{eq:velocity_constraint}.

We solve this problem using a piecewise constant acceleration plan whose segments are aligned to those of the path plan.
This is followed by constant jerk smoothing resulting in a piecewise linear acceleration plan. 
Thus, the velocity plan can also be expressed with a few parameters. 

For segment $i$, given the initial velocity $v_i$ and constant acceleration $a_i$, integrating \eqref{eq:longitudinal_arclength} yields the velocity plan
\begin{equation}\label{eq:const_acc_veclotiy}
    v(s) = \sqrt{v_i^2 + 2 a_i s}.
\end{equation}
To satisfy constraint \eqref{eq:velocity_constraint} and select the highest acceleration possible, the constant acceleration for segment $i$ is given as
\begin{equation}\label{eq:const_acc_plan}
    a_i =\max\bigg\{a_{\rm{min}},\, \min\Big\{a_{\rm{max}},\, \min_{s \in [0,\, s_i]} \frac{1}{2s} \big(\overline{v}^2(s) - v_i^2\big)\Big\}\bigg\}.
\end{equation}
Note that the initial velocities of the second and third segments can be computed according to \eqref{eq:const_acc_veclotiy}:
\begin{equation}\label{eq:initial_velocity}
    v_{1} = \sqrt{v_0^2 + 2 a_0 s_0}, 
    \quad
    v_{2} = \sqrt{v_1^2 + 2 a_1 s_1}.
\end{equation}

The piecewise constant acceleration plan proposed above is discontinuous at the points where the segments meet and violates the jerk constraint. 
Thus, we refine the solution with a smoothing method that interpolates between the segments using the maximum jerk value in \eqref{eq:longitudinal_problem3}. 
For the case ${a_{i-1}>a_i}$, the interpolation occurs within segment ${i-1}$. 
First, we compute the interpolation period $T_{i-1}$ and translate this interval into the traveled distance $S_{i-1}$ assuming the final velocity ${\sqrt{v_{i-1}^2+2a_{i-1}s_{i-1}}}$; cf.~\eqref{eq:const_acc_veclotiy}. 
Then integrating backwards in time with constant jerk $j_{\rm c}$ we obtain
\begin{align}\label{eq:const_jerk_smooth1}
    T_{i-1} &= \frac{a_{i-1}-a_i}{j_c}, 
    \\
    S_{i-1} &= \sqrt{v_{i-1}^2+2a_{i-1}s_{i-1}}T_{i-1} - \frac{1}{2}a_{i-1}T_{i-1}^2 - \frac{2}{3}j_{\rm c}T_{i-1}^3. \nonumber
\end{align}
For ${a_{i-1}\leq a_i}$, the interpolation occurs within segment $i$. 
We compute the corresponding interpolation time and then translate it into the traveled distance by integrating forward in time. This results in
\begin{equation}\label{eq:const_jerk_smooth2}
\begin{split}
    T_i &= \frac{a_i-a_{i-1}}{j_{\rm c}}, 
    \\
    S_i &= \sqrt{v_{i-1}^2+2a_{i-1}s_{i-1}}T_i + \frac{1}{2}a_{i-1}T_i^2 + \frac{1}{6}j_{\rm c}T_i^3.
\end{split}
\end{equation}

That is, for ${a_{0} \leq a_1, a_1 \leq a_2}$, the velocity plan becomes 
\begin{align}\label{eq:vplan1}
    &\texttt{vplan1}(v_0,v_1,v'_2,a_0,a_1,a_2,j_{\rm c},s_0,s_1,s_2,S_1,S_2) 
    \\
    &=
\begin{cases}
    \sqrt{v_0^2+2a_0s}, \quad \text{if } 0 \leq s < s_0,
    \\
    \sqrt{v_1^2+2a_0(s-s_0)+j_{\rm c}(s-s_0)^2}, 
    \\  
    \qquad\qquad\qquad\,\,\, \text{if } s_0\leq s < s_0+S_1,
    \\
    \sqrt{v_1^2+2a_0S_1+j_{\rm c}S_1^2+2a_1(s-s_0-S_1)}, 
    \\  
    \qquad\qquad\qquad\,\,\, \text{if } s_0+S_1 \leq s < s_0+s_1,
    \\
    \sqrt{(v'_2)^2 + 2a_1(s-s_0-s_1)+j_{\rm c}(s-(s_0+s_1))^2}, 
    \\  
    \qquad\qquad\qquad\,\,\,\text{if } s_0+s_1 \leq s < s_0+s_1+S_2,
    \\
    \sqrt{(v'_2)^2+2a_1S_2+j_{\rm c}S_2^2+2a_2(s-s_0-s_1-S_2)}, 
    \\  
    \qquad\qquad\qquad\,\,\, \text{if } s_0+s_1+S_2 \leq s \leq s_\textrm{f},
\end{cases}\nonumber
\end{align}
where
\begin{equation}\label{eq:v_initial_new1}
    v'_2 = \sqrt{v_1^2+2a_0S_1+j_{\rm c}S_1^2+2a_1(s_1-S_1)},
\end{equation}
which can be encoded using 12 parameters. The velocity plans for other cases (${a_{0} > a_1, a_1 > a_2}$; ${a_{0} \leq a_1, a_1 > a_2}$; ${a_{0} > a_1, a_1 \leq a_2}$) are given in Appendix~\ref{sec:appendix_a}. 

\begin{figure}[!t]
\center
\includegraphics[width=3.3 in]{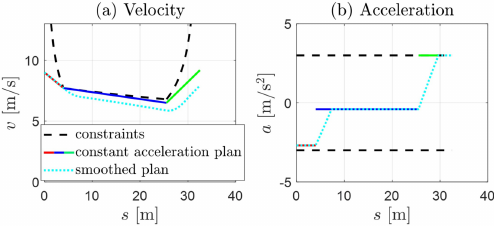}
\caption{Velocity plans using  piece-wise constant acceleration (red, blue, green) and piece-wise linear acceleration (cyan).}
\label{fig:velocity_plan_example}
\vspace{-4mm}
\end{figure}

Fig.~\ref{fig:velocity_plan_example} compares the velocity plans before and after the constant jerk smoothing. 
The velocity profile before smoothing (i.e., with piece-wise constant acceleration) is shown in red, blue, and green, while
the cyan curve depicts the smoothed velocity plan \eqref{eq:vplan1} with piece-wise linear acceleration. 
As indicated by the black dashed curve, both plans satisfy the velocity and acceleration constraints \eqref{eq:longitudinal_problem3} and \eqref{eq:velocity_constraint}, while the smoothing has very little impact on the total travel time.

\section{Implementation}\label{sec:implementation}

In this section, we highlight the advantages of the proposed three-clothoid motion planning algorithm by comparing it with other planning algorithms.
We also describe how the proposed motion plan can be encoded and decoded when sharing it via V2X communication.
Finally, we demonstrate how sharing the motion plans can be utilized for conflict resolution between CAVs.

\subsection{Comparison of planning algorithms}

\begin{figure}[t]
\center
\includegraphics[width=3.3 in]{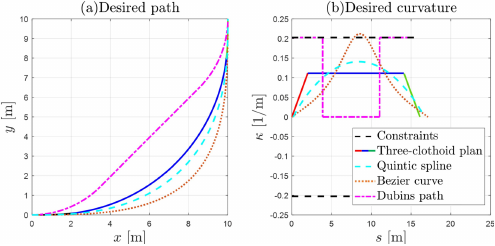}
\includegraphics[width=3.3 in]{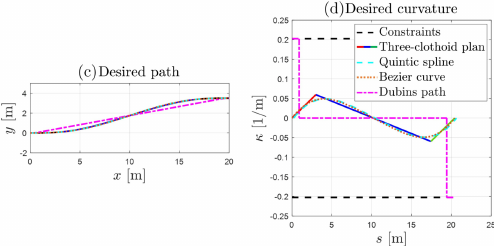}
\includegraphics[width=3.3 in]{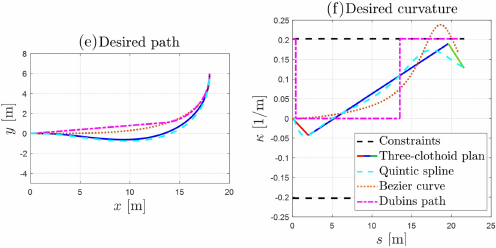}
\caption{Comparison of the results of different planning algorithms.
(a,b) Left turn at an intersection. (c,d) Lane change maneuver. (e,f) Left turn at the parking lot. 
Panels (a,c,e) depict the planned paths while panels (b,d,f) show the curvature plans.}
\label{fig:result_comaprison1}
\vspace{-4mm}
\end{figure}

The motion planning algorithms are demonstrated using three representative urban driving scenarios: (i) left turn at an intersection; (ii) lane change maneuver; and (iii) forward parking into a tight space. 
The first two scenarios are commonly used in the literature while the parking lot scenario is added to test the performance limits of the local planning algorithms. 
For the left turn and the lane change, zero initial and final curvatures are used. 
For the parking lot scenario, nonzero final curvature is set to represent the fact that drivers often use a nonzero steering angle at the end of such maneuvers. 
We compare the three-clothoid planning algorithm with the quintic spline \cite{piazzi2000quintic}, Bezier curve \cite{elbanhawi2015continuous}, and Dubins path \cite{dubins1957curves} methods.
The results are compared quantitatively in Tables~\ref{tb:benchmark_compare} and \ref{tb:computation_compare}, while the designed paths are depicted in Fig.~\ref{fig:result_comaprison1}.

Table~\ref{tb:benchmark_compare} shows that the total length of the path does not vary significantly between different algorithms. 
As proven theoretically \cite{dubins1957curves}, the Dubins path has the shortest length, still, the three-clothoid plan is only up to seven percent longer.
Since the Dubins path is the combination of maximum curvature arcs and a straight line, it does not satisfy $G^2$ continuity making the maximum of the sharpness $\kappa'(s)$ infinite. 
This would lead to jumps in lateral acceleration and poor performance in terms of ride comfort.
The Bezier curve is slightly shorter than the three-clothoid plan, but it does not always satisfy the boundary condition prescribed for the curvature, nor it can guarantee that the curvature $\kappa(s)$ stays below the prescribed limit 0.2; see Fig.~\ref{fig:result_comaprison1}(b),(f).

\begin{table*}[!t]
    \centering
        \caption{Comparison between different local planning methods with various metrics. }
    \begin{tabular}{|c|c|c|c|c|c|c|}
        \hline
        Scenario & Planning method & Total length [m] & Total time [sec] & $\max{\kappa(s)}$ [$\textrm{m}^{-1}$]& $\max{\kappa'(s)}$ [$\textrm{m}^{-2}$] & Boundary condition
        \\
        \hline\hline
        Left turn & Three-clothoid plan & 16.2 & 3.90 & 0.12 & 0.072  & Exactly satisfied
        \\
        & Quintic spline & 16.6 & 4.00 & 0.14 & 0.038 & Exactly satisfied
        \\
        & Bezier curve & 17.1 & 4.21 & 0.21 & 0.042 & Curvature violated
        \\
        & Dubins path & 15.1 & 4.21 & 0.20 & $\infty$ & Exactly satisfied
        \\
        \hline
        Lane change & Three-clothoid plan & 20.4 & 3.84 & 0.060 & 0.020  & Exactly satisfied
        \\
        & Quintic spline & 20.6 & 3.80 & 0.049 & 0.026 & Exactly satisfied
        \\
        & Bezier curve & 20.3 & 3.74 & 0.052 & 0.0056 & Exactly satisfied
        \\
        & Dubins path & 20.3 & 4.21 & 0.20 & $\infty$ & Exactly satisfied
        \\
        \hline
        Parking lot turn & Three-clothoid plan & 21.6 & 4.97 & 0.19 & 0.031  & Exactly satisfied
        \\
        & Quintic spline & 22.1 & 5.01 & 0.17 & 0.061 & Exactly satisfied
        \\
        & Bezier curve & 20.9 & 4.78 & 0.24 & 0.045 & Curvature violated
        \\
        & Dubins path & 20.8 & 5.17 & 0.20 & $\infty$ & Exactly satisfied
        \\
        \hline
    \end{tabular}
    \label{tb:benchmark_compare}
    \vspace{-4mm}
\end{table*}

\begin{table}[!t]
    \centering
        \caption{Average computation time to generate a feasible path.}
    \begin{tabular}{|c|c|}
        \hline
        Planning method & Computation time [sec]\\
        \hline\hline
        Three-clothoid plan & 0.02
        \\
        \hline
        Quintic spline & 0.05
        \\
        \hline
        Bezier curve & 0.08
        \\
        \hline
        Dubins path & 0.005\\
        \hline
    \end{tabular}
    \label{tb:computation_compare}
\end{table}

To be able to evaluate the travel time of the other path planning methods, we used a minimum-time velocity optimizer \cite{consolini2016linear}, which considered only the longitudinal acceleration constraint in \eqref{eq:constraint} and the lateral acceleration constraint \eqref{eq:ride_comfort_constraint}, while ignoring the jerk limit and steering rate constraints. 
In contrast, the travel time for the three-clothoid plan is derived using the piece-wise linear acceleration plan proposed above, which satisfies all constraints and boundary conditions. 
The results in Table~\ref{tb:benchmark_compare} demonstrate that the proposed three-clothoid plan gives only a mild trade-off, if any, in travel time compared to the best of the other methods. 

Table~\ref{tb:computation_compare} compares the computation time between different planning methods and shows that three-clothoid plans can be generated faster than Bezier curves and quintic splines. 
The mean computation time required to generate a feasible path (if there exists one) is given under randomly chosen boundary conditions. 
Specifically, we randomly pick three waypoints while using zero curvature boundary conditions. 
We remark that both the quintic spline and the Bezier curve require parameter searches, which include identifying a range of candidate parameters and evaluating the feasibility of the paths. 
In the case of the quintic spline, one needs to solve a constrained nonlinear optimization problem \cite{piazzi2000quintic,ghilardelli2013path}.
For the Bezier curve the challenge is to locate the control points such that the resultant path is feasible \cite{elhoseny2018bezier,chen2013lane,bae2019lane,yang2010analytical}. 
As opposed to these, the proposed three-clothoid plan has two tunable parameters: the lengths of the first and the last clothoid $s_0$ and $s_2$.  
These parameters, as all other parameters in the three-clothoid plan, have clear physical meaning and they can be tuned to maximize passenger comfort.

\subsection{Encoding and decoding of motion plans}

Once the path plan and the velocity plan are generated, the solution is parameterized as in \eqref{eq:curv} and \eqref{eq:vplan1}. 
Since \eqref{eq:curv} only encodes the curvature, the initial configuration ${(x_0, y_0, \psi_0)}$ should be added. 
Thus, the whole trajectory can be encoded using 19 parameters $(x_0, y_0, \psi_0, s_0, s_1, s_2, \kappa_0, \kappa_1, \kappa_2, \kappa'_1, v_0, v'_1, v'_2, a_0, a_1, a_2,j_{\rm c}$, $S_0, S_1)$.
Note that we did not include the sharpnesses $\kappa'_0$ and $\kappa'_2$ since these can be obtained from $\kappa_0$, $\kappa_1$, $\kappa_2$, and $\kappa'_1$ using \eqref{eq:curvature_continuity}. 
The encoded plan can be shared with nearby vehicles and the infrastructure via V2X communication.
Once receiving the encoded plan, it can be decoded up to the required precision/resolution while the sampling length can be chosen freely. 
Since the velocity plan aligns with the clothoid segments, the desired position is coupled with the desired speed, which is favorable for trajectory tracking. 

We remark that to decode the path onto ${(x, y)}$ position, numerical integration of the curvature function \eqref{eq:curv} from the initial configuration is needed since there is no analytical solution of Fresnel integrals \eqref{eq:Fresnel}. 
There are different methods of numerical evaluation of these definite integrals including Taylor approximation and adaptive quadrature methods. 
For instance, to fully decode a path plan for a left turn maneuver in an intersection with 0.1 m sampling length of the traveled distance, when considering the waypoint ${(\Delta x, \Delta y, \Delta \psi) = (14.5\,\rm{m}, 21.5\,\rm{m}, \frac{\pi}{2}})$, it takes 0.064 seconds with adaptive quadrature method embedded in the Matlab function ``integral".

\subsection{Collision checking and conflict resolution}

\begin{figure}[!t]
\centering
\includegraphics[width=3.4 in]{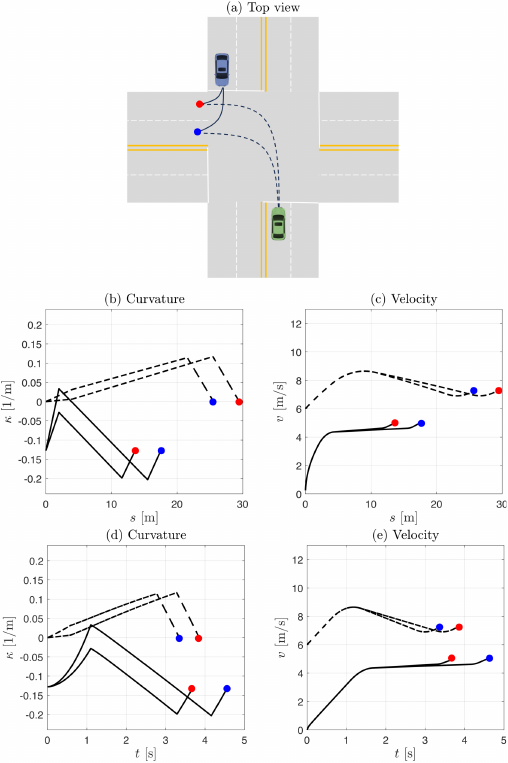}
\caption{Demonstration of shareable motion planning and collision checking for conflict resolution at an intersections.
(a) Driving scenario where the green vehicle is making a left turn and the blue vehicle is making a right turn. 
(b)-(c) Curvature and velocity profiles as a function of the arclength. 
(d)-(e) Curvature and velocity profiles as a function of the time.}
\label{fig:intersection_demo}
\vspace{-4mm}
\end{figure}

Thanks to the simplicity of the proposed motion planning algorithm, it is possible to implement collision checking between plans shared by different connected automated vehicles. 
This can be executed either by the CAVs themselves or by a central server which receives the plans via V2X connectivity. 
Integrating collision checking with motion planning empowers the system to effectively handle potential conflicts and promote smooth interaction among road participants. 

Potential collisions can be detected by finding the existence of intersecting points of clothoids in the ${(x,y)}$ plane; see Fig.~\ref{fig:intersection_demo}(a). 
Using the formalism defined in \eqref{eq:G2_problem_int} and \eqref{eq:Fresnel}, and setting the initial configurations to be ${(x_{0,{\rm a}},y_{0,{\rm a}},\psi_{0,{\rm a}})}$ and ${(x_{0,{\rm b}},y_{0,{\rm b}},\psi_{0,{\rm b}})}$, the intersections of two clothoids is given by the algebraic equations
\begin{equation}\label{eq:curve_intersection}
\begin{split}
    x_{0,{\rm a}} + s_{\rm a} C(\kappa'_{\rm a} s_{\rm a}^2,\kappa_{\rm a} s_{\rm a},\psi_{0,{\rm a}}) 
    &= x_{0,{\rm b}} + s_{\rm b} C(\kappa'_{\rm b} s_{\rm b}^2,\kappa_{\rm b} s_{\rm b},\psi_{0,{\rm a}}),
    \\
    y_{0,{\rm a}} + s_{\rm a} S(\kappa'_{\rm a} s_{\rm a}^2,\kappa_{\rm a} s_{\rm a},\psi_{0,{\rm a}}) 
    &= y_{0,{\rm b}} + s_{\rm b} S(\kappa'_{\rm b} s_{\rm b}^2,\kappa_{\rm b} s_{\rm b},\psi_{0,{\rm b}}).  
\end{split}  
\end{equation}
These can be solved numerically for the arclenghts $s_{\rm a}$ and $s_{\rm b}$ using Newton's method.
Since no discretization of the paths are required, the location of the potential conflict point can be determined with high accuracy.
Collision checking between two clothoids can be generalized to collision checking between two three-clothoid paths with small increase in complexity.

Fig.~\ref{fig:intersection_demo} illustrates how shareable motion planning with collision-checking can be used to avoid conflicts at an intersection. 
In panel (a), the green vehicle aims to execute an unprotected left turn, while the blue vehicle is making a right turn. 
Consider that each vehicle have the option to designate either the red or blue point as their target point. 
The corresponding motion plans, i.e., the curvature and the velocity, are depicted in panels (b) and (c) as a function of the arclength. 
The same plans are given as a function of time in panels (d) and (e).
By sharing the plans via V2X communication and performing collision checking enables the vehicles to detect an avoid potential conflicts.
In particular, if the left turning vehicle selects the blue point and the right turning vehicle selects the red point then no conflict occurs as the paths do not intersect.
If both vehicles select the red point then conflict occurs as they reach this point at approximately the same time. 
If both of them selects the blue point then conflict can be avoided as they reach that point with about 1.3 second difference; see panel (e). 
Finally, if the left turning vehicle selects the red point and the right turning vehicle selects the blue point then the paths intersect and they reach the intersection point with less than a second difference which may be too short to avoid a potential conflict.

In order to improve the precision of collision checking, we can incorporate swept volumes in the analysis. 
The swept volume encompasses the collective future trajectories of all points on the vehicle's body, creating a closed set within the workspace. 
In Appendix~\ref{sec:appendix_b}, we prove that, using a clothoid path and a rectangular bounding box,
the boundary of the swept volume is given by the union of traces of the end points of the rear axle and the corners of the front bumper. 
As the traces of these points are also given by clothoids the above collision checking can be generalized for the swept volumes.

\section{Conclusion}\label{sec:conclusion}

Continuous local motion plans are proposed for connected automated vehicles which can be encoded and regenerated using 19 parameters. 
A three-clothoid-based method was used for planning smooth paths with bounded curvature. 
A corresponding velocity planner generated smooth velocity profiles which were aligned with the paths and obeyed the acceleration and jerk limits. 
Each of the 19 parameters within the planned trajectory holds distinct and interpretable physical meaning.
Such plans can be shared via V2X communication with limited bandwidth and can be decoded by the recipient in a fast and accurate manner, enabling collision checking with high accuracy.
Compared to other interpolating spline-based techniques, the proposed method offers advantages with its light computational and communication load, as well as its smoothness, exactness, compactness, tunability, and interpretability. 
Furthermore, the proposed local motion planner can be tailored for specific road geometries including unusual intersections or curved obstacle courses, and directly integrated with a wide range of higher-level global motion planners that specify boundary conditions. 
The comprehensive discussion on the overall path planning, combining global and local planners, will be the focus of future research.

\bibliographystyle{IEEEtran}
\bibliography{main}

\begin{thebibliography}{10}
\providecommand{\url}[1]{#1}
\csname url@samestyle\endcsname
\providecommand{\newblock}{\relax}
\providecommand{\bibinfo}[2]{#2}
\providecommand{\BIBentrySTDinterwordspacing}{\spaceskip=0pt\relax}
\providecommand{\BIBentryALTinterwordstretchfactor}{4}
\providecommand{\BIBentryALTinterwordspacing}{\spaceskip=\fontdimen2\font plus
\BIBentryALTinterwordstretchfactor\fontdimen3\font minus \fontdimen4\font\relax}
\providecommand{\BIBforeignlanguage}[2]{{%
\expandafter\ifx\csname l@#1\endcsname\relax
\typeout{** WARNING: IEEEtran.bst: No hyphenation pattern has been}%
\typeout{** loaded for the language `#1'. Using the pattern for}%
\typeout{** the default language instead.}%
\else
\language=\csname l@#1\endcsname
\fi
#2}}
\providecommand{\BIBdecl}{\relax}
\BIBdecl

\bibitem{ersal2020connected}
T.~Ersal, I.~Kolmanovsky, N.~Masoud, N.~Ozay, J.~Scruggs, R.~Vasudevan, and G.~Orosz, ``Connected and automated road vehicles: state of the art and future challenges,'' \emph{Vehicle System Dynamics}, vol.~58, no.~5, pp. 672--704, 2020.

\bibitem{wubing2022}
W.~B. Qin, Y.~Zhang, D.~Tak\'acs, G.~St\'ep\'an, and G.~Orosz, ``Nonholonomic dynamics and control of road vehicles: moving toward automation,'' \emph{Nonlinear Dynamics}, vol. 110, no.~3, pp. 1959--2004, 2022.

\bibitem{reif1979complexity}
J.~H. Reif, ``Complexity of the mover's problem and generalizations,'' in \emph{20th Annual Symposium on Foundations of Computer Science}.\hskip 1em plus 0.5em minus 0.4em\relax IEEE, 1979, pp. 421--427.

\bibitem{bertsekas1997nonlinear}
D.~P. Bertsekas, ``Nonlinear programming,'' \emph{Journal of the Operational Research Society}, vol.~48, no.~3, pp. 334--334, 1997.

\bibitem{karaman2011sampling}
S.~Karaman and E.~Frazzoli, ``Sampling-based algorithms for optimal motion planning,'' \emph{The International Journal of Robotics Research}, vol.~30, no.~7, pp. 846--894, 2011.

\bibitem{schmerling2015optimal}
E.~Schmerling, L.~Janson, and M.~Pavone, ``Optimal sampling-based motion planning under differential constraints: the driftless case,'' in \emph{IEEE International Conference on Robotics and Automation}.\hskip 1em plus 0.5em minus 0.4em\relax IEEE, 2015, pp. 2368--2375.

\bibitem{webb2013kinodynamic}
D.~J. Webb and J.~Van~den Berg, ``Kinodynamic {RRT}*: Asymptotically optimal motion planning for robots with linear dynamics,'' in \emph{IEEE International Conference on Robotics and Automation}.\hskip 1em plus 0.5em minus 0.4em\relax IEEE, 2013, pp. 5054--5061.

\bibitem{karaman2013sampling}
S.~Karaman and E.~Frazzoli, ``Sampling-based optimal motion planning for non-holonomic dynamical systems,'' in \emph{IEEE International Conference on Robotics and Automation}.\hskip 1em plus 0.5em minus 0.4em\relax IEEE, 2013, pp. 5041--5047.

\bibitem{lavalle2006planning}
S.~M. LaValle, \emph{Planning Algorithms}.\hskip 1em plus 0.5em minus 0.4em\relax Cambridge University Press, 2006.

\bibitem{gonzalez2016review}
D.~Gonz{\'a}lez, J.~P{\'e}rez, V.~Milan{\'e}s, and F.~Nashashibi, ``A review of motion planning techniques for automated vehicles,'' \emph{IEEE Transactions on Intelligent Transportation Systems}, vol.~17, no.~4, pp. 1135--1145, 2015.

\bibitem{fleury1995primitives}
S.~Fleury, P.~Soueres, J.-P. Laumond, and R.~Chatila, ``Primitives for smoothing mobile robot trajectories,'' \emph{IEEE Transactions on Robotics and Automation}, vol.~11, no.~3, pp. 441--448, 1995.

\bibitem{kellynagy2003}
A.~Kelly and B.~Nagy, ``Reactive nonholonomic trajectory generation via parametric optimal control,'' \emph{The International Journal of Robotics Research}, vol.~22, no. 7-8, pp. 583--601, 2003.

\bibitem{botros2022tunable}
A.~Botros and S.~L. Smith, ``Tunable trajectory planner using {G}$^3$ curves,'' \emph{IEEE Transactions on Intelligent Vehicles}, vol.~7, no.~2, pp. 273--285, 2022.

\bibitem{bertolazzi2018g2}
E.~Bertolazzi and M.~Frego, ``On the $\textrm{G}^2$ hermite interpolation problem with clothoids,'' \emph{Journal of Computational and Applied Mathematics}, vol. 341, pp. 99--116, 2018.

\bibitem{orosz2016connected}
G.~Orosz, ``Connected cruise control: modelling, delay effects, and nonlinear behaviour,'' \emph{Vehicle System Dynamics}, vol.~54, no.~8, pp. 1147--1176, 2016.

\bibitem{jin2018connected}
J.~I. Ge and G.~Orosz, ``Connected cruise control among human-driven vehicles: Experiment-based parameter estimation and optimal control design,'' \emph{Transportation Research Part C}, vol.~95, pp. 445--459, 2018.

\bibitem{naus2010string}
G.~J. Naus, R.~P. Vugts, J.~Ploeg, M.~J. van~de Molengraft, and M.~Steinbuch, ``String-stable {CACC} design and experimental validation: A frequency-domain approach,'' \emph{IEEE Transactions on Vehicular Technology}, vol.~59, no.~9, pp. 4268--4279, 2010.

\bibitem{he2019fuel}
C.~R. He, J.~I. Ge, and G.~Orosz, ``Fuel efficient connected cruise control for heavy-duty trucks in real traffic,'' \emph{IEEE Transactions on Control Systems Technology}, vol.~28, no.~6, pp. 2474--2481, 2019.

\bibitem{liu2020freeway}
H.~Liu, S.~E. Shladover, X.-Y. Lu, and X.~Kan, ``Freeway vehicle fuel efficiency improvement via cooperative adaptive cruise control,'' \emph{Journal of Intelligent Transportation Systems}, pp. 1--13, 2020.

\bibitem{shen2021saving}
M.~Shen, T.~G. Moln{\'a}r, C.~R. He, A.~H. Bell, M.~Hunkler, D.~Oppermann, R.~Zukouski, J.~Yan, and G.~Orosz, ``Saving energy with delayed information in connected vehicle systems,'' in \emph{American Control Conference (ACC)}.\hskip 1em plus 0.5em minus 0.4em\relax IEEE, 2021, pp. 1625--1630.

\bibitem{wang2023intent1}
H.~M. Wang, S.~S. Avedisov, T.~G. Moln\'ar, A.~H. Sakr, O.~Altintas, and G.~Orosz, ``Conflict analysis for cooperative maneuvering with status and intent sharing via {V2X} communication,'' \emph{IEEE Transactions on Intelligent Vehicles}, vol.~8, no.~2, pp. 1105--1118, 2023.

\bibitem{wang2023intent2}
H.~M. Wang, S.~S. Avedisov, O.~Altintas, and G.~Orosz, ``Multi-vehicle conflict management with status and intent sharing under time delays,'' \emph{IEEE Transactions on Intelligent Vehicles}, vol.~8, no.~2, pp. 1624--1637, 2023.

\bibitem{van2021cooperative}
R.~van Hoek, J.~Ploeg, and H.~Nijmeijer, ``Cooperative driving of automated vehicles using {B-}splines for trajectory planning,'' \emph{IEEE Transactions on Intelligent Vehicles}, vol.~6, no.~3, pp. 594--604, 2021.

\bibitem{hult2020optimisation}
R.~Hult, M.~Zanon, S.~Gros, H.~Wymeersch, and P.~Falcone, ``Optimisation-based coordination of connected, automated vehicles at intersections,'' \emph{Vehicle System Dynamics}, vol.~58, no.~5, pp. 726--747, 2020.

\bibitem{chalaki2021priority}
B.~Chalaki and A.~A. Malikopoulos, ``A priority-aware replanning and resequencing framework for coordination of connected and automated vehicles,'' \emph{IEEE Control Systems Letters}, vol.~6, pp. 1772--1777, 2021.

\bibitem{blanco2015tp}
J.~L. Blanco, M.~Bellone, and A.~Gimenez-Fernandez, ``{TP}-space {RRT}--kinematic path planning of non-holonomic any-shape vehicles,'' \emph{International Journal of Advanced Robotic Systems}, vol.~12, no.~5, p.~55, 2015.

\bibitem{dubins1957curves}
L.~E. Dubins, ``On curves of minimal length with a constraint on average curvature, and with prescribed initial and terminal positions and tangents,'' \emph{American Journal of Mathematics}, vol.~79, no.~3, pp. 497--516, 1957.

\bibitem{hwan2013optimal}
J.~Hwan~Jeon, R.~V. Cowlagi, S.~C. Peters, S.~Karaman, E.~Frazzoli, P.~Tsiotras, and K.~Iagnemma, ``Optimal motion planning with the half-car dynamical model for autonomous high-speed driving,'' in \emph{American Control Conference (ACC)}.\hskip 1em plus 0.5em minus 0.4em\relax IEEE, 2013, pp. 188--193.

\bibitem{elbanhawi2015continuous}
M.~Elbanhawi, M.~Simic, and R.~N. Jazar, ``Continuous path smoothing for car-like robots using {B}-spline curves,'' \emph{Journal of Intelligent \& Robotic Systems}, vol.~80, no.~1, pp. 23--56, 2015.

\bibitem{simba2016real}
K.~R. Simba, N.~Uchiyama, and S.~Sano, ``Real-time smooth trajectory generation for nonholonomic mobile robots using {B}ezier curves,'' \emph{Robotics and Computer-Integrated Manufacturing}, vol.~41, pp. 31--42, 2016.

\bibitem{gomez2012optimal}
M.~G{\'o}mez, R.~Gonz{\'a}lez, T.~Mart{\'\i}nez-Mar{\'\i}n, D.~Meziat, and S.~S{\'a}nchez, ``Optimal motion planning by reinforcement learning in autonomous mobile vehicles,'' \emph{Robotica}, vol.~30, no.~2, pp. 159--170, 2012.

\bibitem{sivashangaran2021intelligent}
S.~Sivashangaran and M.~Zheng, ``Intelligent autonomous navigation of car-like unmanned ground vehicle via deep reinforcement learning,'' \emph{IFAC-PapersOnLine}, vol.~54, no.~20, pp. 218--225, 2021.

\bibitem{aradi2020survey}
S.~Aradi, ``Survey of deep reinforcement learning for motion planning of autonomous vehicles,'' \emph{IEEE Transactions on Intelligent Transportation Systems}, vol.~23, no.~2, pp. 740--759, 2020.

\bibitem{piazzi2000quintic}
A.~Piazzi and C.~G.~L. Bianco, ``Quintic {$G^2$}-splines for trajectory planning of autonomous vehicles,'' in \emph{IEEE Intelligent Vehicles Symposium (IV)}.\hskip 1em plus 0.5em minus 0.4em\relax IEEE, 2000, pp. 198--203.

\bibitem{ghilardelli2013path}
F.~Ghilardelli, G.~Lini, and A.~Piazzi, ``Path generation using ${\eta}^4$-splines for a truck and trailer vehicle,'' \emph{IEEE Transactions on Automation Science and Engineering}, vol.~11, no.~1, pp. 187--203, 2013.

\bibitem{qian2016motion}
X.~Qian, I.~Navarro, A.~de~La~Fortelle, and F.~Moutarde, ``Motion planning for urban autonomous driving using b{\'e}zier curves and {MPC},'' in \emph{19th IEEE Intelligent Transportation Systems Conference (ITSC)}.\hskip 1em plus 0.5em minus 0.4em\relax IEEE, 2016, pp. 826--833.

\bibitem{elhoseny2018bezier}
M.~Elhoseny, A.~Tharwat, and A.~E. Hassanien, ``Bezier curve based path planning in a dynamic field using modified genetic algorithm,'' \emph{Journal of Computational Science}, vol.~25, pp. 339--350, 2018.

\bibitem{huang2021personalized}
C.~Huang, H.~Huang, P.~Hang, H.~Gao, J.~Wu, Z.~Huang, and C.~Lv, ``Personalized trajectory planning and control of lane-change maneuvers for autonomous driving,'' \emph{IEEE Transactions on Vehicular Technology}, vol.~70, no.~6, pp. 5511--5523, 2021.

\bibitem{sharma2019survey}
O.~Sharma, N.~Sahoo, and N.~Puhan, ``A survey on smooth path generation techniques for nonholonomic autonomous vehicle systems,'' in \emph{45th Annual Conference of the IEEE Industrial Electronics Society}, vol.~1.\hskip 1em plus 0.5em minus 0.4em\relax IEEE, 2019, pp. 5167--5172.

\bibitem{fraichard2004reeds}
T.~Fraichard and A.~Scheuer, ``From reeds and shepp's to continuous-curvature paths,'' \emph{IEEE Transactions on Robotics}, vol.~20, no.~6, pp. 1025--1035, 2004.

\bibitem{wilde2009computing}
D.~K. Wilde, ``Computing clothoid segments for trajectory generation,'' in \emph{IEEE/RSJ International Conference on Intelligent Robots and Systems}.\hskip 1em plus 0.5em minus 0.4em\relax IEEE, 2009, pp. 2440--2445.

\bibitem{tian2021continuous}
Y.~Tian, Z.~Chen, C.~Xue, Y.~Sun, and B.~Liang, ``Continuous curvature turns based method for least maximum curvature path generation of autonomous vehicle,'' in \emph{47th Annual Conference of the IEEE Industrial Electronics Society}.\hskip 1em plus 0.5em minus 0.4em\relax IEEE, 2021, pp. 1--6.

\bibitem{marzbani2015better}
H.~Marzbani, R.~N. Jazar, and M.~Fard, ``Better road design using clothoids,'' in \emph{Sustainable Automotive Technologies 2014}.\hskip 1em plus 0.5em minus 0.4em\relax Springer, 2015, pp. 25--40.

\bibitem{banzhaf2018g}
H.~Banzhaf, N.~Berinpanathan, D.~Nienh{\"u}ser, and J.~M. Z{\"o}llner, ``From {$G^2$} to {$G^3$} continuity: Continuous curvature rate steering functions for sampling-based nonholonomic motion planning,'' in \emph{IEEE Intelligent Vehicles Symposium (IV)}.\hskip 1em plus 0.5em minus 0.4em\relax IEEE, 2018, pp. 326--333.

\bibitem{kong2015kinematic}
J.~Kong, M.~Pfeiffer, G.~Schildbach, and F.~Borrelli, ``Kinematic and dynamic vehicle models for autonomous driving control design,'' in \emph{IEEE Intelligent Vehicles Symposium (IV)}.\hskip 1em plus 0.5em minus 0.4em\relax IEEE, 2015, pp. 1094--1099.

\bibitem{mi2023UAV}
T.~Mi, D.~Tak\'acs, H.~Liu, and G.~Orosz, ``Capturing the true bounding boxes: vehicle kinematic data extraction using unmanned aerial vehicles,'' \emph{submitted}, 2023.

\bibitem{bachrathy2012bisection}
D.~Bachrathy and G.~St{\'e}p{\'a}n, ``Bisection method in higher dimensions and the efficiency number,'' \emph{Periodica Polytechnica Mechanical Engineering}, vol.~56, no.~2, pp. 81--86, 2012.

\bibitem{consolini2016linear}
L.~Consolini, M.~Locatelli, A.~Minari, and A.~Piazzi, ``A linear-time algorithm for minimum-time velocity planning of autonomous vehicles,'' in \emph{24th Mediterranean Conference on Control and Automation}.\hskip 1em plus 0.5em minus 0.4em\relax IEEE, 2016, pp. 490--495.

\bibitem{chen2013lane}
J.~Chen, P.~Zhao, T.~Mei, and H.~Liang, ``Lane change path planning based on piecewise {B}ezier curve for autonomous vehicle,'' in \emph{IEEE International Conference on Vehicular Electronics and Safety}.\hskip 1em plus 0.5em minus 0.4em\relax IEEE, 2013, pp. 17--22.

\bibitem{bae2019lane}
I.~Bae, J.~H. Kim, J.~Moon, and S.~Kim, ``Lane change maneuver based on {B}ezier curve providing comfort experience for autonomous vehicle users,'' in \emph{22nd IEEE Intelligent Transportation Systems Conference (ITSC)}.\hskip 1em plus 0.5em minus 0.4em\relax IEEE, 2019, pp. 2272--2277.

\bibitem{yang2010analytical}
K.~Yang and S.~Sukkarieh, ``An analytical continuous-curvature path-smoothing algorithm,'' \emph{IEEE Transactions on Robotics}, vol.~26, no.~3, pp. 561--568, 2010.

\end{thebibliography}

\appendices

\section{Velocity plan formulation} \label{sec:appendix_a}

For ${a_{0} > a_1, a_1 > a_2}$, the velocity plan becomes 
\begin{align}\label{eq:vplan2}
    &\texttt{vplan2}(v_0,v'_1,v'_2,a_0,a_1,a_2,j_{\rm c},s_0,s_1,s_2,S_0,S_1) 
    \\
    &=
\begin{cases}
    \sqrt{v_0^2+2a_0s}, \quad \text{if } 0 \leq s < s_0-S_0,
    \\
    \sqrt{ 
        \begin{aligned}
            v_0^2&+2a_0(s_0-S_0)+2a_0(s-s_0+S_0)\\
            &-j_{\rm c}(s^2-(s_0-S_0)^2)+2j_{\rm c}s_0(s_0-S_0-s)
        \end{aligned}
        }, 
    \\  
    \qquad\qquad\qquad\,\,\, \text{if } s_0-S_0 \leq s < s_0,
    \\
    \sqrt{(v'_1)^2+2a_1(s-s_1)}, \quad \text{if } s_0 \leq s < s_0+s_1-S_1,
    \\
    \sqrt{
        \begin{aligned}
            (v'_1)^2&+2a_1(s_1-S_1)+2a_1(s-s_0-s_1+S_1)\\
            &-j_{\rm c}(s^2-(s_0+s_1-S_1)^2)\\
            &+2j_{\rm c}(s_0+s_1)(s_0+s_1-S_1-s)
        \end{aligned}
        },
    \\  
    \qquad\qquad\qquad\,\,\,\text{if } s_0+s_1-S_1 \leq s < s_0+s_1,
    \\
    \sqrt{(v'_2)^2+2a_2(s-s_0-s_1)},
    \\
    \qquad\qquad\qquad\,\,\,\text{if } s_0+s_1 \leq s \leq s_\textrm{f},
\end{cases}\nonumber
\end{align}
where
\begin{equation}\label{eq:v_initial_new2}
    v'_1 = \sqrt{v_0^2+2a_0s_0-j_{\rm c}S_0^2}, 
    \quad
    v'_2 = \sqrt{(v'_1)^2+2a_1s_1-j_{\rm c}S_1^2}.
\end{equation}

For ${a_{0} \leq a_1, a_1 > a_2}$, the velocity plan becomes 
\begin{align}\label{eq:vplan3}
    &\texttt{vplan3}(v_0,v_1,v'_2,a_0,a_1,a_2,j_{\rm c},s_0,s_1,s_2,S_1,S_2) 
    \\
    &=
\begin{cases}
    \sqrt{v_0^2+2a_0s}, \quad \text{if } 0 \leq s < s_0,
    \\
    \sqrt{v_1^2+2a_0(s-s_0)+j_{\rm c}(s-s_0)^2}, 
    \\  
    \qquad\qquad\qquad\,\,\, \text{if } s_0\leq s < s_0+S_1,
    \\
    \sqrt{v_1^2+2a_0S_1+j_{\rm c}S_1^2+2a_1(s-s_0-S_1)}, 
    \\  
    \qquad\qquad\qquad\,\,\, \text{if } s_0+S_1 \leq s < s_0+s_1-S_2,
    \\
    \sqrt{
        \begin{aligned}
            (v'_2)^2&+2a_1(s-s_0-s_1+S_2)\\
            &-j_{\rm c}(s-(s_0+s_1-S_2))^2
        \end{aligned}
        },
    \\  
    \qquad\qquad\qquad\,\,\,\text{if } s_0+s_1-S_2 \leq s < s_0+s_1,
    \\
    \sqrt{(v'_2)^2+2a_1S_2-j_{\rm c}S_2^2+2a_2s}, 
    \\  
    \qquad\qquad\qquad\,\,\, \text{if } s_0+s_1 \leq s \leq s_\textrm{f},
\end{cases}\nonumber
\end{align}
where
\begin{equation}\label{eq:v_initial_new3}
    v'_2 = \sqrt{v_1^2+2a_0S_1+j_{\rm c}S_1^2+2a_1(s_1-S_1-S_2)}.
\end{equation}

For ${a_{0} > a_1, a_1 \leq a_2}$, the velocity plan becomes 
\begin{align}\label{eq:vplan4}
    &\texttt{vplan4}(v_0,v'_1,v'_2,a_0,a_1,a_2,j_{\rm c},s_0,s_1,s_2,S_0,S_2) 
    \\
    &=
\begin{cases}
    \sqrt{v_0^2+2a_0s}, \quad \text{if } 0 \leq s < s_0-S_0,
    \\
    \sqrt{ 
        \begin{aligned}
            v_0^2&+2a_0(s_0-S_0)+2a_0(s-s_0+S_0)\\
            &-j_{\rm c}(s^2-(s_0-S_0)^2)+2j_{\rm c}s_0(s_0-S_0-s)
        \end{aligned}
        }, 
    \\  
    \qquad\qquad\qquad\,\,\, \text{if } s_0-S_0 \leq s < s_0,
    \\
    \sqrt{(v'_1)^2+2a_1s}, \quad \text{if } s_0 \leq s < s_0+s_1,
    \\
    \sqrt{
        \begin{aligned}
            (v'_2)^2&+2a_1(s-s_0-s_1)+j_{\rm c}(s-(s_0+s_1))^2
        \end{aligned}
    }, 
    \\  
    \qquad\qquad\qquad\,\,\,\text{if } s_0+s_1 \leq s < s_0+s_1+S_2,
    \\
    \sqrt{(v'_2)^2+2a_1S_2+j_{\rm c}S_2^2+2a_2(s-s_0-s_1-S_2)}, 
    \\  
    \qquad\qquad\qquad\,\,\, \text{if } s_0+s_1+S_2 \leq s \leq s_\textrm{f},
\end{cases} \nonumber
\end{align}
where
\begin{equation}\label{eq:v_initial_new4}
    v'_1 = \sqrt{v_0^2+2a_0s_0-j_{\rm c}S_0^2}, 
    \quad
    v'_2 = \sqrt{(v'_1)^2+2a_1s_1}.
\end{equation}

\section{Swept volume boundary of a clothoid path}\label{sec:appendix_b}

\begin{figure}[!t]
\center
\includegraphics[width=2.0 in]{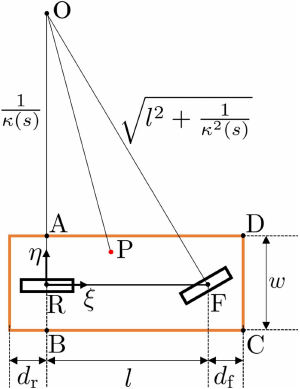}
\caption{Rectangular approximation of the occupied region in the workspace at the initial state. 
An arbitrary point P within the rectangle can be parameterized as ${(\xi,\eta)}$ while considering the center of the rear axle R as the origin. 
The center O of the osculating circle is depicted as the intersection between two line segments which are normal to the rear wheel and to the front wheel.}
\label{fig:vehicle_square}
\vspace{-4mm}
\end{figure}

To construct the boundary of the swept volume one may compute the traces of the points A, B, C, and D in Fig~\ref{fig:vehicle_square}: 
\begin{equation}\label{eq:rear_axle1}
\begin{split}
    x_{\rm A}(s) & = \hat{x}_i +  s C\big(\kappa_i' s^2, \hat{\kappa}_i s, \hat{\psi}_i \big) - \textstyle\frac{w}{2} \sin\psi(s)  
    \\
    y_{\rm A}(s) & = \hat{y}_i + s S\big(\kappa_i' s^2, \hat{\kappa}_i s, \hat{\psi}_i \big) + \textstyle\frac{w}{2} \cos\psi(s)
\end{split}  
\end{equation}
\begin{equation}\label{eq:rear_axle2}
\begin{split}
    x_{\rm B}(s) & = \hat{x}_i + s C\big(\kappa_i' s^2, \hat{\kappa}_i s, \hat{\psi}_i \big) + \textstyle\frac{w}{2} \sin\psi(s),  
    \\
    y_{\rm B}(s) & = \hat{y}_i + s S\big(\kappa_i' s^2, \hat{\kappa}_i s, \hat{\psi}_i \big) - \textstyle\frac{w}{2} \cos\psi(s),
\end{split}  
\end{equation}
\begin{equation}\label{eq:front_bumper1}
\begin{split}
    x_{\rm C}(s)= & \hat{x}_i + s C\big(\kappa_i' s^2, \hat{\kappa}_i s, \hat{\psi}_i \big) 
    \\
    & \quad + \textstyle\frac{w}{2} \sin\psi(s) + (l+d_{\textrm{f}}) \cos\psi(s), 
    \\
    y_{\rm C}(s)= & \hat{y}_i + s S\big(\kappa_i' s^2, \hat{\kappa}_i s, \hat{\psi}_i \big)  
    \\
    & \quad - \textstyle\frac{w}{2} \cos\psi(s) + (l+d_{\textrm{f}}) \sin\psi(s) ,
\end{split}  
\end{equation}
\begin{equation}\label{eq:front_bumper2}
\begin{split}
    x_{\rm D}(s)= & \hat{x}_i + s C\big(\kappa_i' s^2, \hat{\kappa}_i s, \hat{\psi}_i \big) 
    \\
    & \quad - \textstyle\frac{w}{2} \sin\psi(s) + (l+d_{\textrm{f}}) \cos\psi(s), 
    \\
    y_{\rm D}(s)= & \hat{y}_i + s S\big(\kappa_i' s^2, \hat{\kappa}_i s, \hat{\psi}_i \big)  
    \\
    & \quad + \textstyle\frac{w}{2} \cos\psi(s) + (l+d_{\textrm{f}}) \sin\psi(s),
\end{split}  
\end{equation}
where ${i=0,1,3}$ refers to the individual clothoid. 
Here $\kappa_i'$ denotes the sharpness of the $i$-th clothoid, $\hat{\kappa}_i$ denotes its initial curvature, and
\begin{equation}\label{eq:psi}
\psi(s)=\kappa'_i \frac{s^2}{2} + \hat{\kappa}_i s + \hat{\psi}_i,
\end{equation}
where $\hat{\psi}_i$ is the initial yaw angle.
For the our three-clothoid design we have ${\hat{\kappa}_0 = \kappa_0}$, ${\hat{\kappa}_1 = \kappa_1 - \kappa_1' \frac{s_1}{2}}$, ${\hat{\kappa}_2 = \kappa_2 - \kappa_2' s_2}$ (cf.~\eqref{eq:curv}) and ${\hat{x}_0=0}$, ${\hat{y}_0=0}$, ${\hat{\psi}_0=0}$ (cf.~\eqref{eq:G2_problem_BC}), while the expressions for $\hat{x}_1, \hat{y}_1, \hat{\psi}_1, \hat{x}_2, \hat{y}_2, \hat{\psi}_2$ are omitted for simplicity. 

\noindent \textbf{Proposition 1.} \textit{The boundary of the swept volume of a rectangle that follows a clothoid path consists of the rectangle for the initial and final configuration, the traces of the endpoints  A and B of the rear axle, and the traces of the corners C and D of the front bumper.}

\noindent \textbf{Definition 1}: The occupied region of an object in the 2D workspace $\mathcal{W}$ is defined as the semialgebraic set. 

\noindent \textbf{Definition 2}: The swept volume $SW$ of a moving object is the union of future traces of the occupied region of the object when following a pre-defined path. 

\noindent \textbf{Definition 3}: The boundary of the swept volume $\partial SW$ is defined as the boundary of the closed set ${SW \subset \mathcal{W}}$. 

\noindent \textbf{Claim 1}: The boundary of the swept volume of the rectangle (vehicle's body) can be approximated by the traces of the of four edges of the rectangle. 

\noindent \textbf{Proof of Claim 1}: At each time instance, part of the boundary of the swept volume is given by the edges of the rectangle. 
The interior points are inside the rectangle by definition. 
As the rectangle progresses along the path, every interior point is enclosed by the union of the four edges. 

Consider the coordinate system defined in Fig.~\ref{fig:vehicle_square}, where the center of the rear axle R is the origin and $\xi$ and $\eta$ point to the longitudinal and lateral directions, respectively. 

\noindent \textbf{Claim 2}: For a clothoid segment ${\kappa(s) = \hat{\kappa}_i+\kappa'_i s}$ with non-negative initial curvature (${\hat{\kappa}_i \geq 0}$) and non-negative sharpness (${\kappa'_i \geq 0}$), the trace of an arbitrary point along the left edge, i.e., $(\xi,\frac{w}{2}), -d_{{\rm r}}\leq \xi \leq l+d_{{\rm f}}$, is inside the trace of ${\rm A}$ located at ${(0,\frac{w}{2})}$. 

\begin{figure}[!t]
\center
\includegraphics[width=86mm]{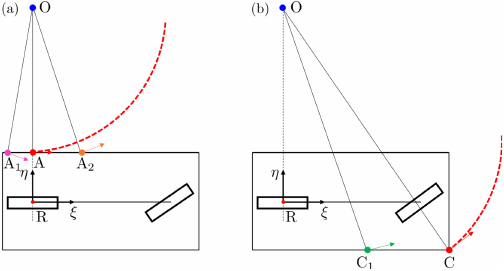}
\caption{Traces of points on the left and right edges. 
(a) illustrates the motion of the points ${\rm A}$, ${\rm A}_1$, and ${\rm A}_2$ on the left edge. 
(b) illustrates the motion of points ${\rm C}$ and ${\rm C}_1$ on the right edge. 
${\rm O}$ is the instantaneous center of rotation.}
\label{fig:left_right_edge_case1}
\vspace{-4mm}
\end{figure}

\noindent \textbf{Proof of Claim 2}: In Fig.~\ref{fig:left_right_edge_case1}(a), two points on the left edge are depicted: ${\rm A}_{1}$ is located at ${(\xi,\frac{w}{2})}$, ${-d_{\rm r} \leq \xi \leq 0}$ while ${\rm A}_{2}$ is located at ${(\xi,\frac{w}{2})}$, ${0 \leq \xi \leq l+d_{\rm f}}$. 

Given the initial curvature ${\hat{\kappa}_i \geq 0}$ and the sharpness ${\kappa'_i \geq 0}$, two things should be proven.  
\\
\textbf{Claim 2.1}: The trace of ${\rm A}_{1}$ is inside the left edge of the rectangle at the initial position or inside the trace of ${\rm A}$. 
\\ 
\textbf{Claim 2.2}: The trace of ${\rm A}_{2}$ is inside the trace of ${\rm A}$. 

\begin{figure}[!t]
\center
\includegraphics[width=82mm]{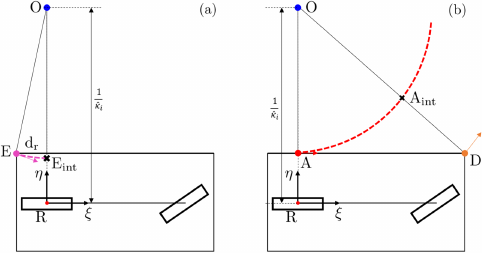}
\caption{Intersecting points ${\rm E}_{\rm int}$ and ${\rm A}_{\rm int}$ needed for the proof of Claim~2.}
\label{fig:left_edge}
\end{figure}

The point ${\rm A}_1$ is the furthest from ${\rm A}$ when it coincides with point ${\rm E}$, the left corner of the rear bumper. 
This point is located at ${(-d_{{\rm r}}, \frac{w}{2})}$ in the ${(\xi,\eta)}$ coordinate system; see  Fig.~\ref{fig:left_edge}(b).
Thus, the corresponding trajectory is given by 
\begin{equation}\label{eq:rear_bumper2}
\begin{split}
    x_{\rm E}(s)= & s C\big( \kappa'_i s^2, \hat{\kappa}_i s, 0 \big) 
    - \textstyle\frac{w}{2} \sin\psi(s) + d_{\textrm{r}} \cos\psi(s), 
    \\
    y_{\rm E}(s)= & s S\big( \kappa'_i s^2, \hat{\kappa}_i s, 0 \big)  
    + \textstyle\frac{w}{2} \cos\psi(s) + d_{\textrm{r}} \sin\psi(s),
\end{split}  
\end{equation}
where ${i=0,1,2}$ refer to the individual clothoid, such that 
$\kappa_i'$ denotes the sharpness of the $i$-th clothoid,  $\hat{\kappa}_i$ denotes its initial curvature, and
\begin{equation}\label{eq:psi2}
\psi(s)=\kappa'_i \frac{s^2}{2} + \hat{\kappa}_i s .
\end{equation}
Here without loss of generality the initial position of R and the initial yaw angle is considered to be zero; cf.~\eqref{eq:rear_axle1},\eqref{eq:rear_axle2},\eqref{eq:front_bumper1},\eqref{eq:front_bumper2},\eqref{eq:psi}.

\begin{figure}[!t]
\center
\includegraphics[width=2.4 in]{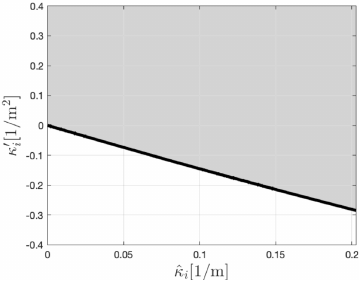}
\caption{Gray shading indicates the values of initial curvature $\hat{\kappa}_i$ and sharpness $\kappa'_i$ for which ${\eta_{{\rm int},1}\leq\frac{w}{2}}$, that is, $\textrm{E}_{\textrm{int}}$ is below ${\rm A}$. 
The vehicle parameters are ${(l+d_{{\rm f}}) = 3.8~{\rm m}}$, ${d_{\rm r} = 1~{\rm m}}$, ${w = 1.9~{\rm m}}$.}
\label{fig:left_rear_result1}
\vspace{-4mm}
\end{figure}

To prove \textbf{Claim 2.1}, one should show that the coordinate $\eta$ of the intersecting point ${\rm E}_{{\rm int}}$ is smaller than or equal to $\frac{w}{2}$ along the line $\overline{{\rm O} {\rm R}}$ as depicted in Fig.~\ref{fig:left_edge}. 
Based on the path equation \eqref{eq:rear_bumper2}, we can write 
\begin{equation}\label{eq:E_int1x}
    0 = s C(\kappa'_i s^2,\hat{\kappa}_i s,0) - \textstyle\frac{w}{2} \sin\psi(s) - d_{\rm r} \cos(\psi(s),
\end{equation}
where $\psi(s)$ is given by \eqref{eq:psi2}.
Once solving this equation for the arclength $\bar{s}$, we should determine the value
\begin{equation}\label{eq:E_int1y}
\begin{split}
    \eta_{{\rm int},1} = \bar{s} S(\kappa'_i \bar{s}^2,\kappa_i \bar{s},0)  + \textstyle\frac{w}{2} \cos\psi(\bar{s}) - d_{{\rm r}} \sin\psi(\bar{s}).
\end{split}  
\end{equation}
When choosing the vehicle parameters $l, d_{{\rm f}}, d_{{\rm r}}, w$,
it can be shown numerically that within the shaded domain in the ${(\hat{\kappa}_i, \kappa'_i)}$-plane in Fig.~\ref{fig:left_rear_result1} we have ${\eta_{{\rm int},1}\leq\frac{w}{2}}$, that is, $\textrm{E}_{\textrm{int}}$ is below ${\rm A}$. 
This proves \textbf{Claim 2.1}.

The point ${\rm A_2}$ is the furthest from ${\rm A}$ when it coincides with ${\rm D}$, the left corner of the front bumper which is located at ${(l+d_{{\rm f}}, \frac{w}{2})}$ at the ${(\xi,\eta)}$ coordinate system. 
The trajectory of A in Fig.~\ref{fig:left_edge}(b) is given by 
\begin{equation}\label{eq:rear_axle3}
\begin{split}
    x_{\rm A}(s)= & s C\big( \kappa'_i s^2, \hat{\kappa}_i s, 0 \big) 
    - \textstyle\frac{w}{2} \sin\psi(s),
    \\
    y_{\rm A}(s)= & s S\big( \kappa'_i s^2, \hat{\kappa}_i s, 0 \big)  
    + \textstyle\frac{w}{2} \cos\psi(s),
\end{split}  
\end{equation}
where ${i=0,1,2}$ refer to the individual clothoids and $\psi(s)$ is defined in \eqref{eq:psi2}; cf.~\eqref{eq:rear_axle1},\eqref{eq:psi} while considering the initial position of R and the initial yaw angle to be zero.

To prove \textbf{Claim 2.2} we show that the intersecting point ${\rm A_{\rm int}}$ in Fig.~\ref{fig:left_edge}(b) is between ${\rm O}$ and ${\rm D}$ along the line $\overline{{\rm O} {\rm D}}$, that is, the $\eta$ coordinate of ${\rm A_{\rm int}}$ is larger than or equal to $\frac{w}{2}$.
We find the intersecting point ${\rm A}_{\rm{int}}$  by using 
\begin{equation}\label{eq:E_int2x}
    \frac{\frac{1}{\hat{\kappa}_i}-\frac{w}{2}}{ l+d_{\rm f}}x_{\rm A}(s)+y_{\rm A}(s)=\frac{1}{\hat{\kappa}_i}.
\end{equation}
where $x_{\rm A}(s)$ and  $y_{\rm A}(s)$ are defined in \eqref{eq:rear_axle3}.
Once solving this for $\bar{\bar{s}}$ we can determine the value 
\begin{equation}
  \eta_{{\rm int},2}= \bar{\bar{s}} S\big( \kappa'_i \bar{\bar{s}}^2, \hat{\kappa}_i \bar{\bar{s}}, 0 \big)  
    + \textstyle\frac{w}{2} \cos\psi(\bar{\bar{s}}).
\end{equation}
When choosing the vehicle parameters $l, d_{{\rm f}}, d_{{\rm r}}, w$, it can be numerically shown that for non-negative values of $\hat{\kappa}_i$ and $\kappa'_i$ 
we have ${\eta_{{\rm int},2} \geq \frac{w}{2}}$, that is, the length of $\overline{{\rm OA_{int}}}$ is shorter than $\overline{{\rm O} {\rm D}}$. This proves \textbf{Claim 2.2}.

\noindent \textbf{Claim 3}: For a clothoid segment ${\kappa(s) = \hat{\kappa}_i+\kappa'_i s}$ with non-negative initial curvature (${\hat{\kappa}_i \geq 0}$) and non-negative sharpness (${\kappa'_i \geq 0}$), the traces of arbitrary point on the right edge, i.e., ${(\xi,-\frac{w}{2})}$, ${-d_{{\rm r}}\leq \xi \leq (l+d_{{\rm f}})}$ is inside the trace of ${\rm C}$ located at ${(l+d_\textrm{f},-\frac{w}{2})}$.

\begin{figure}[!t]
\center
\includegraphics[width=35mm]{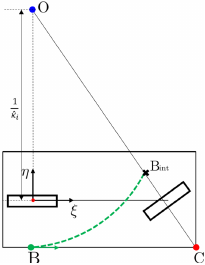}
\caption{Intersecting point  ${\rm B}_{\rm int}$ needed for the proof of Claim~3. }
\label{fig:right_edge}
\vspace{-4mm}
\end{figure}

\noindent \textbf{Proof of Claim 3}: In Fig.~\ref{fig:left_right_edge_case1}(b), a point on the right edge is depicted: ${\rm C}_1$ is located at${(\xi,-\frac{w}{2})}$, ${-d_{{\rm r}} \leq \xi \leq (l+d_{{\rm f}})}$.

Given the initial curvature ${\hat{\kappa}_i \geq 0}$ and the sharpness ${\kappa'_i \geq 0}$, it should be proven that the intersection between the future trace of ${\rm C}_1$ is inside the future trace of ${\rm C}$. 

The point ${\rm C_1}$ is the furthest from ${\rm C}$ when it coincides with ${\rm B}$, the right end of the rear axle. 
This point is located at ${(0, -\frac{w}{2})}$ at the ${(\xi,\eta)}$ coordinate system. 
Thus, the corresponding trajectory is given by 
\begin{equation}\label{eq:right_edge_3}
\begin{split}
    x_{\rm B}(s)= & s C\big( \kappa'_i s^2, \hat{\kappa}_i s, 0 \big) 
    + \textstyle\frac{w}{2} \sin\psi(s),
    \\
    y_{\rm B}(s)= & s S\big( \kappa'_i s^2, \hat{\kappa}_i s, 0 \big)  
    - \textstyle\frac{w}{2} \cos\psi(s),
\end{split}  
\end{equation}
where ${i=0,1,2}$ refer to the individual clothoid and $\psi(s)$ is defined in \eqref{eq:psi2}; cf.~\eqref{eq:rear_axle2},\eqref{eq:psi} while considering the initial position of R and the initial yaw angle to be zero. 

\begin{figure}[!t]
\center
\includegraphics[width=2.4 in]{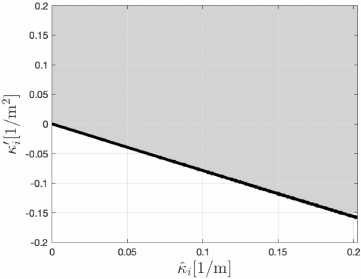}
\caption{Gray shading indicates the values of initial curvature $\hat{\kappa}_i$ and sharpness $\kappa'_i$ for which ${\eta_{{\rm int},3}\geq-\frac{w}{2}}$,, that is, $\textrm{B}_{\textrm{int}}$ is between ${\rm O}$ and ${\rm C}$. 
The vehicle parameters are ${(l+d_{{\rm f}}) = 3.8~{\rm m}}$, ${d_{\rm r} = 1~{\rm m}}$, ${w = 1.9~{\rm m}}$.}
\label{fig:right_front_result1}
\vspace{-4mm}
\end{figure}

To prove \textbf{Claim 3} one should show that the intersection point ${\rm B_{\rm int}}$ in Fig.~\ref{fig:right_edge} is between ${\rm O}$ and ${\rm C}$ along the line $\overline{{\rm O} {\rm C}}$, that is, 
the $\eta$ coordinate of ${\rm B}_{{\rm int}}$ is larger than or equal to $-\frac{w}{2}$ as depicted in Fig.~\ref{fig:right_edge}. 
We find the intersecting point ${\rm B}_{\rm{int}}$ by using 
\begin{equation}\label{eq:B_int2x}
    \frac{\frac{1}{\hat{\kappa}_i}+\frac{w}{2}}{ l+d_{\rm f}}x_{\rm B}(s)+y_{\rm B}(s)=\frac{1}{\hat{\kappa}_i}.
\end{equation}
where $x_{\rm B}(s)$ and  $y_{\rm B}(s)$ are defined in \eqref{eq:right_edge_3}.
Once solving this for $\bar{\bar{\bar{s}}}$ we can determine the value 
\begin{equation}
  \eta_{{\rm int},3}= \bar{\bar{\bar{s}}} S\big( \kappa'_i \bar{\bar{\bar{s}}}^2, \hat{\kappa}_i \bar{\bar{\bar{s}}}, 0 \big)  
    - \textstyle\frac{w}{2} \cos\psi(\bar{\bar{\bar{s}}}).
\end{equation}
When choosing the vehicle parameters $l, d_{{\rm f}}, d_{{\rm r}}, w$, it can be shown numerically that within the shaded domain in the ${(\hat{\kappa}_i, \kappa'_i)}$-plane in Fig.~\ref{fig:right_front_result1} we have ${\eta_{{\rm int},3}} \geq -\frac{w}{2}$, that is, the length of $\overline{{\rm OB_{int}}}$ is shorter than $\overline{{\rm O} {\rm C}}$. 
This proves \textbf{Claim 3}.

\noindent \textbf{Remark 1}: For a clothoid with initial curvature ${\hat{\kappa}_i \leq 0}$ and sharpness ${\kappa'_i \leq 0}$, the future trace of ${\rm D}$ will provide the left boundary in $\partial SW$ and the future trace of ${\rm B}$ will provide the right boundary in $\partial SW$. 

\noindent \textbf{Remark 2}: For a clothoid with initial curvature ${\hat{\kappa}_i \geq 0}$ and sharpness ${\kappa'_i \leq 0}$, the traces of ${\rm D}$ and ${\rm A}$ intersect and the traces of ${\rm C}$ and ${\rm B}$ intersect. 
These intersections can be found for the left boundary by solving the equations
\begin{align}\label{eq:boundary_intersection1}
    & s_{\rm rl} C(\kappa'_i s_{\rm rl}^2,\hat{\kappa}_i s_{\rm rl},\psi_0) - \textstyle\frac{w}{2} \sin\psi(s_{\rm rl})
    \\
    & = s_{\rm fl} C(\kappa'_i s_{\rm fl}^2,\hat{\kappa}_i s_{\rm fl},\psi_0) - \textstyle\frac{w}{2} \sin\psi(s_{\rm fl}) + (l+d_{{\rm f}}) \cos\psi(s_{\rm fl}), \nonumber
    \\
    & s_{\rm rl} S(\kappa'_i s_{\rm rl}^2,\hat{\kappa}_i s_{\rm rl},\psi_0) + \textstyle\frac{w}{2} \cos\psi(s_{\rm rl})\nonumber
    \\
    &=  s_{\rm fl} S(\kappa'_i s_{\rm fl}^2,\hat{\kappa}_i s_{\rm fl},\psi_0) + \textstyle\frac{w}{2} \cos\psi(s_{\rm fl}) + (l+d_{{\rm f}}) \sin\psi(s_{\rm fl}),\nonumber
\end{align}
for $s_{\rm rl}, s_{\rm fl}$. 
For the range ${0 \leq s \leq s_{\rm rl}}$, the trace of the left end ${\rm A}$ of the rear axle expresses the left boundary in $\partial SW$, while for the range ${s \geq s_{\rm fl}}$, the trace of the left corner  ${\rm D}$ of the front bumper expresses the left boundary in $\partial SW$.

Similarly, for the right boundary, we obtain the intersections by solving the equations
\begin{align}\label{eq:boundary_intersection2}
    & s_{\rm rr} C(\kappa'_i s_{\rm rr}^2,\hat{\kappa}_i s_{\rm rr},\psi_0) + \textstyle\frac{w}{2} \sin\psi(s_{\rm rr})
    \\
    &= s_{\rm fr} C(\kappa'_i s_{\rm fr}^2,\hat{\kappa}_i s_{\rm fr},\psi_0) + \textstyle\frac{w}{2} \sin\psi(s_{\rm fr}) + (l+d_{{\rm f}}) \cos\psi(s_{\rm fr}), \nonumber
    \\
    & s_{\rm rr} S(\kappa'_i s_{\rm rr}^2,\hat{\kappa}_i s_{\rm rr},\psi_0) - \textstyle\frac{w}{2} \cos\psi(s_{\rm rr})\nonumber
    \\
    &= s_{\rm fr} S(\kappa'_i s_{\rm fr}^2,\hat{\kappa}_i s_{\rm fr},\psi_0) - \textstyle\frac{w}{2} \cos\psi(s_{\rm fr}) + (l+d_{{\rm f}}) \sin\psi(s_{\rm fr}).\nonumber
\end{align}
for ${s_{\rm rr}, s_{\rm fr}}$.
For the range ${0 \leq s \leq s_{\rm fr}}$, the trace of the right corner ${\rm C}$ of the front bumper expresses the right boundary in $\partial SW$, while for the range $s \geq s_{\rm rr}$, the trace of the right end of the rear axle ${\rm B}$ expresses the right boundary in $\partial SW$. 

\noindent \textbf{Remark 3}: There is a small segment of the swept volume where Proposition 1 does not hold. Specifically, at the initial phase of a clothoid path with positive curvature, the trajectory of the rightmost point of the rear bumper gives the right boundary. Although this variation is negligible, a comprehensive delineation of the swept volume boundaries shall incorporate the initial  traces of the corners of the rear bumper.

\noindent \textbf{Remark 4}: Once the proposition is proven for a single clothoid path, it can be generalized for a continuous three-clothoid path.

\begin{IEEEbiography}
[{\includegraphics[width=1in,height=1.25in,clip,keepaspectratio]{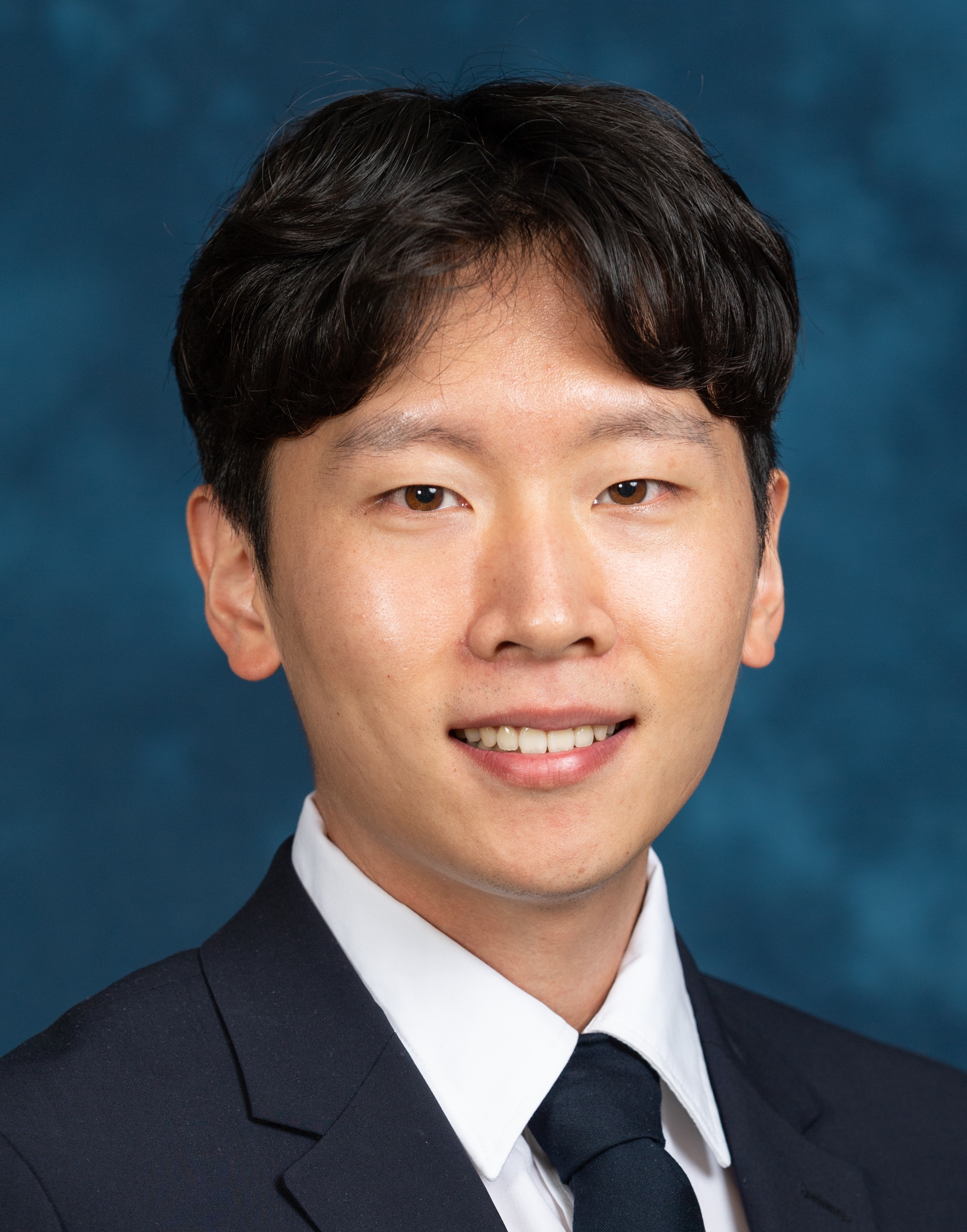}}]{Sanghoon Oh} received the BS degree in Mechanical Engineering from the Seoul National University, Seoul, in 2018. 
He is currently pursuing a PhD degree in Mechanical Engineering at the University of Michigan, Ann Arbor. 
His research interests include vehicle dynamics and control, and motion planning with applications to connected and automated vehicles.
\end{IEEEbiography}

\vspace{-15mm}
\begin{IEEEbiography}
[{\includegraphics[width=1in,height=1.25in,clip,keepaspectratio]{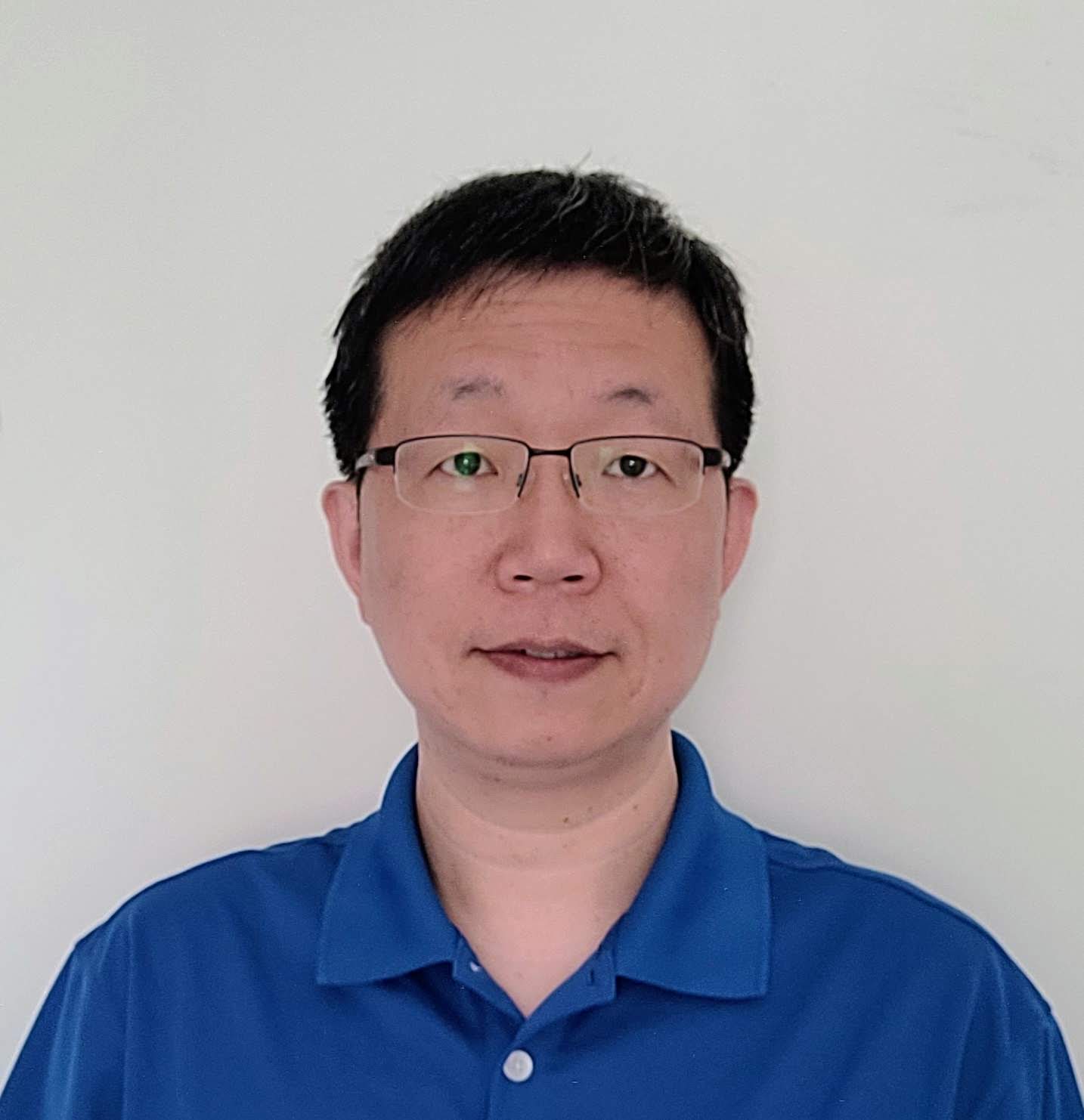}}]{Qi Chen} received the BS and MS degrees from the Zhejiang University, China, in 1997 and 2000, respectively, and the PhD degree in Electrical and Computer Engineering from the Ohio State University, Columbus in 2007. 
He joined the Ford Motor Company in 2016 and currently is a research engineer. 
Before Ford, he was an engineering specialist at Caterpillar Inc. 
\end{IEEEbiography}

\vspace{-15mm}
\begin{IEEEbiography}
[{\includegraphics[width=1in,height=1.25in,clip,keepaspectratio]{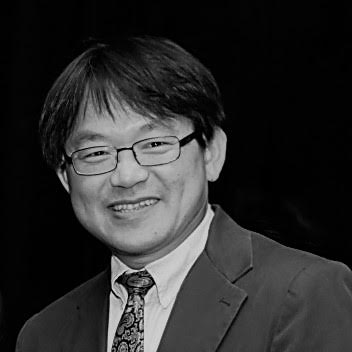}}]{H. Eric Tseng} received the BS degree from the National Taiwan University in 1986, and the MS and PhD degrees in mechanical engineering from the University of California, Berkeley in 1991 and 1994, respectively. 
In 1994, he joined Ford Motor Company. 
At Ford (1994-2022), he had a productive career and retired as a Senior Technical Leader of Controls and Automated Systems in Research and Advanced Engineering. 
Many of his contributed technologies led to production vehicles implementation. 
His technical achievements have been honored with Ford's annual technology award, the Henry Ford Technology Award, on seven occasions. 
Additionally, he was the recipient of the Control Engineering Practice Award from the American Automatic Control Council in 2013. 
He has over 100 U.S. patents and over 160 publications. He is a member of the National Academy of Engineering as of 2021.
\end{IEEEbiography}

\vspace{-15mm}
\begin{IEEEbiography}    [{\includegraphics[width=1in,height=1.25in,clip,keepaspectratio]{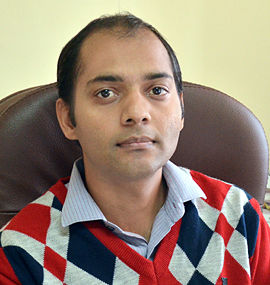}}]{Gaurav Pandey} received his B-Tech from IIT Roorkee in 2006 and completed his PhD from University of Michigan, Ann Arbor in 2013. 
He is currently working as Technical Leader of autonomous vehicles research at Ford Motor Company. 
He leads the team working on developing algorithms for autonomous navigation of vehicles specifically focused on perception. 
Before Ford, Dr Pandey was an Assistant Professor at the Electrical Engineering Department of the Indian Institute of Technology (IIT) Kanpur in India. At IIT Kanpur he was part of two research groups in the Electrical Engineering Department (i) Control and Automation, (ii) Signal Processing and Communication. 
His current research focus is on visual perception for autonomous vehicles and mobile robots using tools from computer vision, machine learning, and information theory. 
\end{IEEEbiography}

\vspace{-15mm}
\begin{IEEEbiography}
[{\includegraphics[width=1in,height=1.25in,clip,keepaspectratio]{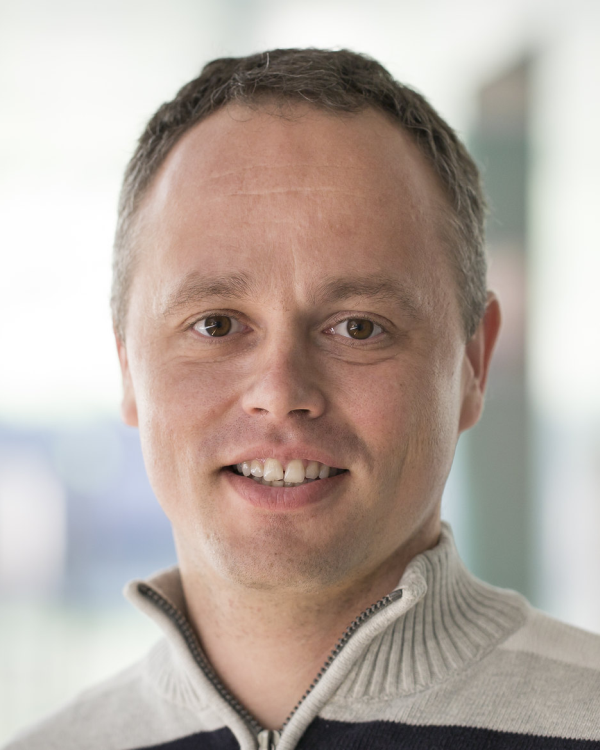}}]{G\'abor Orosz} received the MSc degree in Engineering Physics from the Budapest University of Technology, Hungary, in 2002 and the PhD degree in Engineering Mathematics from the University of Bristol, UK, in 2006. 
He held postdoctoral positions at the University of Exeter, UK, and the University of California, Santa Barbara. 
In 2010, he joined the University of Michigan, Ann Arbor where he is currently an Associate Professor in Mechanical Engineering and Civil and Environmental Engineering. From 2017 to 2018 he was a Visiting Professor in Control and Dynamical Systems at the California Institute of Technology. 
In 2022 he was a Visiting Professor in Applied Mechanics at the Budapest University of Technology. 
His research interests include nonlinear dynamics and control, time delay systems, machine learning, and data-driven systems with applications to connected and automated vehicles, traffic flow, and biological networks.
\end{IEEEbiography}

\end{document}